\documentclass[letterpaper,11pt]{article} 

\usepackage{opticameet3} 

\usepackage{nicefrac}
\usepackage{graphicx}
\usepackage{subcaption}
\usepackage{lipsum}
\usepackage{comment}
\usepackage{amssymb}
\usepackage{amsmath}
\usepackage{bbm}
\usepackage{amsthm}
\usepackage{dsfont}
\usepackage{enumitem}
\usepackage{float}

\usepackage{algorithm}
\usepackage{algpseudocode}
\usepackage{tikz}

\usepackage{tabularx}
\usepackage{booktabs} 

\newtheorem{theorem}{Theorem}[section]

\newtheorem{corollary}[theorem]{Corollary}

\newtheorem{assumption}[theorem]{Assumption}
\newtheorem{proposition}[theorem]{Proposition}

\newcommand\D{\mathcal{D}}
\newcommand\K{\mathcal{K}}
\newcommand\X{\mathcal{X}}
\newcommand\bigO{\mathcal{O}}

\newcommand\R{\mathbb{R}} 
\newcommand\E{\mathbb{E}}
\newcommand\1{\mathbbm{1}}

\newcommand\pr{\mathds{P}}
\newcommand{\norm}[1]{\left\lVert#1\right\rVert}

\DeclareMathOperator*{\argmax}{arg\,max}
\DeclareMathOperator*{\argmin}{arg\,min}
\DeclareMathOperator{\Var}{Var}

\newcounter{relctr} 
\everydisplay\expandafter{\the\everydisplay\setcounter{relctr}{0}} 

\newcommand\labelrel[2]{%
  \begingroup
    \refstepcounter{relctr}%
    \stackrel{\textnormal{(\alph{relctr})}}{\mathstrut{#1}}%
    \originallabel{#2}%
  \endgroup
}
\AtBeginDocument{\let\originallabel\label}

\newcommand\authormark[1]{\textsuperscript{#1}}

\usepackage{amsmath,amssymb}
\usepackage[colorlinks=true,bookmarks=false,citecolor=blue,urlcolor=blue]{hyperref} 

\begin{document}

\title{Pseudo-Bayesian Optimization}


\author{Haoxian Chen,\authormark{1} Henry Lam\authormark{1}}

\address{\authormark{1} Department of Industrial Engineering and Operations Research, Columbia University}

\email{haoxian.chen@columbia.edu, henry.lam@columbia.edu} 

\begin{abstract}
Bayesian Optimization is a popular approach for optimizing expensive black-box functions. Its key idea is to use a surrogate model to approximate the objective and, importantly, quantify the associated uncertainty that allows a sequential search of query points that balance exploitation-exploration. Gaussian process (GP) has been a primary candidate for the surrogate model, thanks to its Bayesian-principled uncertainty quantification power and modeling flexibility. However, its challenges have also spurred an array of alternatives whose convergence properties could be more opaque. Motivated by these, we study in this paper an \emph{axiomatic} framework that elicits the minimal requirements to guarantee black-box optimization convergence that could apply beyond GP-based methods. Moreover, we leverage the design freedom in our framework, which we call \emph{Pseudo-Bayesian Optimization}, to construct empirically superior algorithms. In particular, we show how using simple local regression, and a suitable ``randomized prior" construction to quantify uncertainty, not only guarantees convergence but also consistently outperforms state-of-the-art benchmarks in examples ranging from high-dimensional synthetic experiments to realistic hyperparameter tuning and robotic applications.
\end{abstract}

\section{Introduction}
Bayesian Optimization (BO) is a popular method for the global optimization of black-box functions that are typically multimodal and expensive to evaluate. Its main idea is to model the objective landscape using observed data in order to ``guess" where the global optimum is, rather than following a solution trajectory. One of the earliest works in BO is \cite{jones1998efficient}, initially proposed for automotive and semiconductor designs and motivated by long runtimes of their computer codes. Nowadays, BO has been widely applied in machine learning including hyper-parameter tuning and Auto-ML \cite{JMLR:v23:21-0888}, reinforcement learning \cite{calandra2016bayesian}, robotics \cite{martinez2007active}, experimental design such as A/B testing \cite{chapelle2011empirical}, simulator calibration \cite{sha2020applying, bai2022efficient}, and engineering applications such as environmental monitoring \cite{marchant2012bayesian} and aerospace system design \cite{hebbal2021bayesian}. Extensive overviews are in \cite{brochu2010tutorial, shahriari2015taking, frazier2018tutorial}.

More precisely, a typical BO procedure sequentially searches for next design points to evaluate by updating a surrogate model to predict the objective and, moreover, quantify the associated uncertainty. Both the prediction and the 
uncertainty quantification are important as they work together, via a so-called acquisition function, to balance exploitation-exploration in the evaluation sequence. In this regard, Gaussian process (GP) has been a primary candidate of surrogate model, thanks to its Bayesian-principled uncertainty quantification power and modeling flexibility \cite{jones2001taxonomy, osborne2009gaussian,powell2012optimal,gramacy2020surrogates,ankenman2010stochastic}. However, it has a cubic-order scalability due to the inversion of the Gram kernel matrix in computing the posterior and, despite its principled design, it is conceivable to construct algorithms that could leverage problem structures more efficiently. To this end, there has been a surge of active studies to improve BO. These include the sparsification of GP using pseudo-inputs \cite{seeger2003fast, lawrence2002fast, snelson2005sparse, mcintire2016sparse, hensman2013gaussian, csato2002sparse} or kernel approximation \cite{lazaro2010sparse} to reduce its time complexity, substitution of GP with other surrogates such as random forests \cite{hutter2011sequential} and neural networks \cite{springenberg2016bayesian, snoek2015scalable, perrone2018scalable, white2021bananas}, and most recently the direct modeling of acquisition functions over the search space instead of through a surrogate \cite{bergstra2011algorithms, tiao2021bore, song2022general}. 



Most of the state-of-the-art procedures mentioned above (and expanded further in Section \ref{sec:related works}) are practically motivated. In other words, they are shown to be superior via empirical comparisons, and designed via nicely intuited but ad hoc ideas. At the same time, the theory of BO has been largely confined to GP-based procedures. Among them, algorithmic consistency, i.e., convergence to the true optimum as the algorithm evolves indefinitely, can be attained with broader assumptions and algorithmic structures (but still GP-based; e.g., \cite{locatelli1997bayesian,vazquez2010convergence,frazier2009knowledge,bect2016supermartingale}). Convergence rate results, while giving stronger conclusion, typically require more opaque assumptions, such as those associated with reproducing kernel Hilbert space (RKHS) and the tail decay behaviors of the GPs' gradients or the kernel's spectrum, as well as algorithmic restrictions such as upper confidence bound (UCB) instead of general acquisition functions \cite{srinivas2009gaussian,kandasamy2015high,chowdhury2021no}. In either case, there appears to be a mismatch between theory and practice: On one hand, practically superior algorithms do not have guarantees; on the other hand, BO theory is largely confined to GP and is not broad enough to cover more practical algorithms.

Given the above, our first high-level goal of this paper is to build a theory for exploration-based optimization that encompasses \emph{general} algorithms beyond GP. This makes a step forward to reduce the gap between theory and practice in the realm of BO. To clarify our scope, by theory here we focus on algorithmic consistency -- While this is weaker than rate results, algorithmic consistency is still largely open for procedures beyond GP, and moreover avoids the opaque assumptions and restrictive settings encountered in rate analysis. With this, our second high-level goal is to leverage our theory to locate strongly performing algorithms. Here, by strongly performing we mean algorithms that are empirically competitive against state-of-the-art benchmarks, while simultaneously have theoretical guarantees.

With these goals in mind, our main contribution is an \emph{axiomatic} framework that elicits the minimal requirements on any exploration-based algorithms to guarantee convergence. Specifically, we dissect an exploration-based algorithm into three independent basic ingredients, \emph{surrogate predictor (SP)}, \emph{uncertainty quantifier (UQ)}, and \emph{acquisition function (AF)}. SP aims to provide point predictions at different design points. UQ quantifies the uncertainty of SP and indicates how reliable is the current prediction. AF transforms SP and UQ into the decision on which design point to evaluate next. We derive the axiomatic properties of SP, UQ and AF to attain theoretical convergence. In a nutshell, we call these properties \emph{local consistency}, the \emph{sequential no-empty-ball property}, and the \emph{improvement property} respectively. These basic ingredients, along with their axiomatic properties, appear in GP-based algorithms -- however, and as our key message, GP is not the only approach that exhibits these properties; instead, there are many more algorithms that could lead to similar convergence guarantees. We call our above framework \emph{Pseudo-Bayesian Optimization (PseudoBO)}. This is because, intuitively, we hinge on the Bayesian insight of BO in dissecting convergent procedures, but at the same time, a Bayesian perspective is generally not required for optimization convergence. Essentially, our framework extracts the minimally needed features in BO from the view of exploration-based optimization.

Our PseudoBO framework creates a recipe, with a list of SP, UQ and AF candidates that can be shown to satisfy the respective properties and hence altogether lead to algorithmic convergence. Our next endeavor is to assemble candidates in this recipe that perform competitively against existing benchmarks in empirical experiments. This resulting algorithm would then be one that exhibits both empirical and theoretical advantages over the benchmarks. In particular, we show how combining a simple local regression as SP, a suitably constructed ``randomized prior" as UQ, and expected improvement (EI) as AF, can consistently outperform some state-of-the-art BO benchmarks across examples ranging from high-dimensional synthetic experiments to realistic hyperparameter tuning and robotic applications.

The remainder of this paper is as follows. Section \ref{sec:related works} reviews related works on both practical and theoretical fronts. Section \ref{sec:theory} presents the general theory of PseudoBO. Section \ref{sec:recipe} compiles the recipe that comprises a range of exemplifying ingredients under the PseudoBO framework. Section \ref{sec:implementation} discusses some details in turning our theory into implementation. Section \ref{sec:experiments} presents our empirical performances and comparisons with benchmark approaches. Section \ref{sec:conclusion} concludes our paper and discusses future directions. All proofs and additional numerical details are presented in the Appendix.

\section{Related Works}\label{sec:related works}
We review the existing literature. We categorize it roughly into two parts, one focusing on practical algorithms and implementations (Section \ref{sec:practice lit}) and one on theoretical guarantees (Section \ref{sec:theory lit}).

\subsection{Literature on BO Practical Enhancements}\label{sec:practice lit}
We overview some existing approaches that aim to increase the scalability or efficiency of BO:

\textbf{Sparse GP}.\hspace{1mm} The first approach comprises sparse pseudo-input GP \cite{seeger2003fast, lawrence2002fast, snelson2005sparse, mcintire2016sparse, hensman2013gaussian, csato2002sparse}. The idea is to select $m$ inducing pseudo-inputs by minimizing information loss, with $m$ smaller than number of collected data $n$, to reduce the rank of the involved covariance matrix and subsequently the computational complexity. The second approach is GP with sparse spectrum \cite{lazaro2010sparse}. By Bochner's theorem \cite{bochner2005harmonic}, the kernel can be approximated by random features and thus GP can then be approximated by Bayesian linear regression with finite basis functions of size $m$. In terms of complexity, both approaches reduce the time complexity from $\bigO(n^3)$ to $\bigO(nm^2 + m^3)$ for the posterior distribution updating. A hybrid method of speeding up the posterior sampling by combining the two approximations is \cite{wilson2020efficiently}. However, the involved approximation and hence information loss may cause imprecise surrogate prediction or uncertainty quantification \cite{shahriari2015taking}.

\textbf{Surrogate Substitution}.\hspace{1mm}  Rather than working on scaling GP, there has been a surge of research on replacing BO with other surrogate models. SMAC \cite{hutter2011sequential} leverages random forest regressor and the standard deviation among trees to quantify uncertainty. However, the uncertainty quantification approach in this method, which utilizes bootstrapping, is unstable and can be overly small for unexplored areas. An alternative is to use neural networks for surrogate modeling. DNGO \cite{snoek2015scalable} adopts Bayesian linear regression on top of the the representations learnt by a neural network. Based on this, \cite{perrone2018scalable} proposes ABLR that improves the two-step learning in \cite{snoek2015scalable} to joint Bayesian learning but with additional computational complexity. BOHAMIANN \cite{springenberg2016bayesian} uses a modified Hamiltonian Monte Carlo to improve robustness and scalability of the surrogate model. \cite{white2021bananas} uses neural network ensemble. However, they require retraining the neural network after new data is collected, which is computationally expensive. Moreover, the performance appears sensitive to the network architecture \cite{snoek2015scalable}.

\textbf{Density Ratio Estimation-based Methods}.\hspace{1mm}  Another line of research focuses on directly modeling the acquisition function \cite{bergstra2011algorithms, tiao2021bore, song2022general}. \cite{bergstra2011algorithms} proposes tree-structured Parzen estimator (TPE) by establishing the equivalence between expected improvement (EI) and the relative ratio between two densities, specifically for handling discrete and tree-structured inputs. Recently, BORE \cite{tiao2021bore} is invented to estimate this density ratio directly with likelihood-free inference and LFBO \cite{song2022general} generalizes this framework to model any acquisition function in the form of expected utility. 

\textbf{Prior-data Fitted Networks (PFNs)}.\hspace{1mm}  PFNs, initially proposed in \cite{muller2021transformers}, leverage in-context learning technique and transformer architectures to approximate Bayesian inference. While PFNs significantly speed up the posterior inference, they require substantial pretraining. This pretraining procedure involves repeatedly drawing data points from the prior distribution to effectively learn the probabilistic predictions needed for Bayesian inference. Applications of PFNs to BO include \cite{muller2023pfns4bo} and \cite{rakotoarison2024context}.

\textbf{Other Approaches}.\hspace{1mm} There are a variety of other works to speed up the computation of BO. One way is to leverage batch acquisition function, by proposing points in batch to be evaluated at once. Ensemble Bayesian Optimization \cite{wang2018batched} employs an ensemble of additive GPs as well as a batch acquisition function to scale BO to tens of thousands of observations. Other batch acquisition functions are proposed \cite{chevalier2013fast, wang2020parallel, shah2015parallel, wu2016parallel, gonzalez2016batch}. Another line of works focus on dimension reduction, such as BOCK \cite{oh2018bock} and HeSBO \cite{nayebi2019framework}. Recently, TuRBO \cite{eriksson2019scalable} is proposed to incorporate BO with the trust region method and batch acquisition through Thompson sampling \cite{thompson1933likelihood}. In terms of implementation, GPyTorch \cite{gardner2018gpytorch} scales GP computation to thousands of queries, with conjugate descent to solve linear systems and Lanczos process to approximate the log-determinant. By leveraging hardware development, BoTorch \cite{balandat2020botorch} is invented as the state-of-the-art implementation of BO, speeding up the computation of acquisition with Monte Carlo sampling, sample average approximation (SAA) and computational technologies like auto-differentiation and parallel computation on CPUs and GPUs. 

As explained in our introduction, the above approaches are practically oriented, in the sense that they are empirically attractive but can lack theoretical guarantee even on basic consistency. A contribution from our PseudoBO framework is our investigation of simple and cheap algorithms that can perform competitively against these benchmarks, while at the same time exhibit consistency guarantee. Furthermore, we also note that PseudoBO is not meant to separate from the above works, in the sense that these developed tools can be combined with PseudoBO to offer better performance or accelerate the computation even more. 





\subsection{Literature on BO Theory}\label{sec:theory lit}
We divide the study of BO algorithms with theoretical guarantees into the investigation of consistency and more elaborate convergence rate analysis.

\textbf{Consistency}.\hspace{1mm}  \cite{locatelli1997bayesian} analyzes the consistency of BO with one-dimensional GP (in particular, Brownian motion) with EI. This is expanded upon in \cite{vazquez2010convergence}, which extends the consistency of EI to multi-dimensional settings, assuming the GP is stationary with the inverse of its spectral density exhibiting at most polynomial growth. \cite{frazier2011consistency} studies the consistency of Bayesian information collection strategies on noisy observations following exponential family distributions, and applies their sufficient conditions to optimal computing budget allocation and knowledge gradient (KG). Later, \cite{bect2016supermartingale} generalizes the consistency to acquisitions of both EI and KG for any GP with continuous sample paths. Besides, consistency results have also been established in various non-standard settings. For instance, \cite{toscano2018bayesian} considers objectives that are sums or integrals of multiple black-box functions. \cite{astudillo2021bayesian} considers objectives with inner network structures, where each node represents a black-box function with deterministic outputs. More generally, \cite{astudillo2019bayesian} studies composite objectives where a black-box function is composed with a function that has an explicitly known form. In the realm of simulation optimization, BO is applied to optimize expectation-form objective functions, and the simulator can generate aleatory uncertainties via possibly common random numbers, e.g., \cite{pearce2022bayesian, xie2016bayesian}. Additionally, \cite{ungredda2022bayesian} considers black-box functions with input parameters that are unknown but observed via external data (i.e., the so-called input uncertainty in the stochastic simulation literature, e.g., \cite{barton2022input,corlu2020stochastic}). They propose Bayesian Information Collection and Optimization (BICO), which balances the trade-off between simulation and real data collection.

\textbf{Convergence Rate Analysis}.\hspace{1mm} This is cast commonly in the form of cumulative regret or simple regret.  Cumulative regret measures the cumulative difference between the attained objective value and the oracle best objective value attained by the unknown ground-truth optimal solution. Simple regret, on the other hand, measures the difference between the best objective value obtained so far and the oracle best objective value.

Assuming the underlying true objective resides in the RKHS space of linear, RBF, or Mátern kernel, \cite{srinivas2009gaussian} derives the first cumulative regret that scales linearly with the maximum information gain -- a kernel-dependent complexity measure -- and the square root of the total steps, in the noisy setting for UCB acquisition. However, this bound does not guarantee sub-linear regrets for practical kernels like Matérn. More recently, \cite{whitehouse2024sublinear} refines the analysis of UCB for objectives in RKHS induced by Matérn, ensuring sub-linear regrets. Similarly, \cite{valko2013finite} proposes KernelUCB, a kernelized UCB algorithm, improving the cumulative regret bound by reducing the linear dependence on maximum information gain to square root, though the algorithm is less practical. On the lower bound end, \cite{scarlett2017lower} derives the first sub-linear lower bound for any algorithm with Matérn and RBF kernels. There are algorithms focusing more on exploration: \cite{vakili2021optimal} proposes an algorithm called Maximum Variance Reduction (MVR), attaining an optimal order simple regret proportional to maximum information gain and the inverse of the total steps for objectives with bounded norm in an RKHS. \cite{salgia2023random} studies random exploration, proposing an algorithm with region shrinking, which achieves order-optimal cumulative regret in both noiseless and noisy settings. Additionally, \cite{bull2011convergence} analyzes the regret for the EGO algorithm in deterministic settings. \cite{liu2023global} establishes a regret bound for UCB when the objective belongs to a parametric family of functions, and \cite{li2023convergence} gives a concentration bound especially when using $\epsilon$-greedy procedures in radial basis function interpolations. Notably, all these bounds are developed for specific algorithms, except for \cite{scarlett2017lower}. Lastly, we also note the work of \cite{tuo2022uncertainty} that studies confidence region construction for optimal solution and objective value by using the solution iterates from BO.

It is worth mentioning the extensive literature on multi-armed bandit problems, including for instance \cite{kaufmann2012thompson, russo2014learning} that develop Bayesian regret bounds for so-called Thompson sampling. However, the discrete or linear structures of these problems appear different from continuous-space black-box optimization. In particular, their uncertainty typically arises from data noise, while in our setting there can be zero noise from data and the uncertainty comes solely from the lack of knowledge about the unexplored portion of the objective function.

With the exceptions of \cite{liu2023global} and \cite{li2023convergence}, all works above on analyzing BO focus on GP-based procedures. In contrast, PseudoBO derives an axiomatic dissection on consistent exploration-based algorithms beyond GP. In this way, we expand the scope of algorithms above that satisfy consistency. On the other hand, we do not offer rate results as in the second line of works above. However, we note that the theory on consistency already appears largely open beyond GP, and moreover avoids the more opaque assumptions typically needed for sharper rate analyses.

\section{Theory of Pseudo-Bayesian Optimization}\label{sec:theory}

Suppose we are interested in solving the optimization problem
$    \max_{x \in \X} f(x)$,
where the objective function $f:\X\to\R$ is unknown and $\X \subset \R^d$ is the decision space. The observations are deterministic. To introduce PseudoBO, we start with a simple but general algorithmic framework as follows. We optimize $f$ by using sequential function evaluations, where selecting which point to evaluate next is guided by some \emph{evaluation worthiness (EW)} measure, say $W_n(x; \D_n)$. Here, $n$ is the step index in the procedure, and $\D_n=\{(x_1,f(x_1)),\ldots,(x_n,f(x_n))\}$ denotes the collected data up to the $n$-th step. For convenience, we also denote $X_n=\{x_1,\ldots,x_n\}$. In PseudoBO, at each step $n$ we solve $\max_{x\in \X} W_n(x; \D_n)$ and the solution $x_{n+1}$ is set as the next point to evaluate. Thus, in summary, the PseudoBO algorithm is: 

\emph{\ \ For each iteration $n = 1,2,...,T$:} 

\emph{\ \ \ \ \ \ Evaluate $f(x_n)$;}

\emph{\ \ \ \ \ \ Update $W_n(x; \D_n)$;}

\emph{\ \ \ \ \ \  Set $x_{n+1}\in\text{argmax}_{x\in \X}W_n(x;\D_n)$.}
\\


The estimated optimal solution at any step $n$ is $\hat x_n^*\in\text{argmax}_{x\in X_n}f(x)$ which gives the maximum evaluated function value so far.


\subsection{Basic Algorithmic Consistency}

Our first result characterizes the algorithmic consistency of PseudoBO.
We denote $Z^*=\max_{x \in \X} f(x)$ as the optimal value of the target problem, $\Delta(x,S)=\min_{y\in S}\|x-y\|$ as the set distance from $x \in \X$ to the set $S \subset \X$, and $E_f(S)=\{(x,f(x)):x\in S\}$ as the set of evaluated pairs $(x,f(x))$ for all $x\in S$. 

\begin{assumption}[Sequential no-empty-ball property]
$W_n(\cdot\ ;\ \cdot)$ satisfies the following:
\begin{enumerate}[leftmargin=*]

\item For any $x\in\X$ and finite-cardinality set sequence $S_n\subset\X$, if $\inf_{n} \Delta(x, S_{n}) > 0$, then $\liminf_{n\to \infty}W_n(x; A_{n}) > 0$ where $A_n=E_{f}(S_n)$.

\item For any convergent sequence $x_n\in\X$, i.e., $x_n\to x'$ for some $x'\in\X$, we have $W_n(x_n;A_{n-1}\cup \D_{n-1})\to0$, where $\D_{n-1}=\{(x_1,f(x_1)),\ldots,(x_{n-1},f(x_{n-1}))\}$ and $A_n=E_{f}(S_n)$ for any finite-cardinality set sequence $S_n\subset\X$.
\end{enumerate}
\label{basic assumptions}
\end{assumption}

Roughly speaking, part 1 of Assumption \ref{basic assumptions} stipulates that as along as there is no infinitesimally close evaluated point in the neighborhood of $x$, the EW of $x$ is positive. Part 2 plays a converse role to state that if $x$ entails an approaching sequence, then the EW of $x$ would shrink to 0, and this is true with or without any additional data represented by the set sequence $\{A_n\}$. The finite cardinality condition on the considered set sequences is imposed since we focus on realistic algorithms that can evaluate only a finite number of points, even though conceptually the assumption can be relaxed to include any set sequences (but in this case the EW for specific examples need to be properly defined). Note that Assumption \ref{basic assumptions} is purely about the function $W_n$ and its interaction with $f$, and does not assume anything about the optimization procedure. We call Assumption \ref{basic assumptions} the \emph{sequential no-empty-ball (SNEB)} property, where the no-empty-ball (NEB) notion follows from \cite{vazquez2010convergence} and indicates that a zero value of $W_n$ at a point $x$ means any ball surrounding $x$ must contain some evaluated points in the past data, and vice versa. It is ``sequential" because part 2 of the assumption modifies the original NEB property to consider the value of $W_n$ at the sequence $x_n$ instead of its limit $x'$ as in \cite{vazquez2010convergence}. Moreover, we also allow $W_n$ to be step-dependent, i.e., depend on step $n$. The former modification facilitates the argument of our basic guarantee, while the latter is useful to apply to some important examples as we will see in the sequel.



We introduce our first theoretical result: As long as the EW $W$ satisfies the SNEB property, PseudoBO asymptotically reaches the true optimal value, or in other words it is \emph{algorithmically consistent}.
\begin{theorem}[Algorithmic consistency of PseudoBO]
Suppose EW $W_n$ satisfies Assumption \ref{basic assumptions} and $\X$ is compact. Then:
\begin{itemize}
\item $\X$ is eventually populated by the PseudoBO iterates, i.e., for any $x\in\X$, we have $\inf_n\Delta(x,X_n)=0$ for $X_n=\{x_1,\ldots,x_n\}$ where $x_n$ is the $n$-th output iterate in the PseudoBO algorithm. 
\item Consequently, if furthermore $f$ is continuous, then PseudoBO is algorithmically consistent, i.e., the estimated optimal solution $\hat x_n^*\in\text{argmax}_{x\in X_n}f(x)$ satisfies $f(\hat x_n^*)\to Z^*$ as $n\to\infty$.
\end{itemize}\label{main basic}
\end{theorem}

Theorem \ref{main basic} reveals the exploration-based nature of PseudoBO: It achieves convergence by populating the search space. More precisely, the first part of the theorem concludes eventual popularization, which means that any $x\in\X$ has arbitrarily close evaluated points from the PseudoBO iterates eventually. By selecting the historically best evaluated point, this popularization then turns into asymptotic convergence to the optimal value, which is the second part of the theorem. Here, while EW can incorporate many sources of information, a requirement is that it must contain information about the local popularity to guide us in this space popularization. In reality, we would like to make guesses and evaluate at points that are likely close to the best (exploitation), but also be cautious about missing out other potentially good regions (exploration). Accounting for this tradeoff requires a more specialized framework that contains ingredients to handle this issue more explicitly. We will describe these ingredients in the next subsection.

\subsection{A More Specialized Framework}\label{sec:specialized}

We consider a more specialized version of PseudoBO that materializes EW via three ingredients: \emph{surrogate predictor (SP)}, \emph{uncertainty quantifier (UQ)} and \emph{acquisition function (AF)}. As discussed earlier, these ingredients appear in GP-based algorithms in BO but could be designed substantially more generally. Each of these ingredients needs to satisfy its own basic, independent, assumption, which we call \emph{local consistency}, \emph{SNEB} (introduced before) and the \emph{improvement property} respectively.

\begin{assumption}[Local consistency of SP]
The SP $\hat f(\cdot;\cdot): \X\times E_{f}(\X)\to\mathbb R$ satisfies that for any convergent sequence $\{x_n\}\subset \X$, i.e., $x_n\to x'$ for some $x'\in \X$, we have $\hat f(x_n;A_{n-1}\cup \D_{n-1})\to f(x')$, where $\D_{n-1}=\{(x_1,f(x_1)),\ldots,(x_{n-1},f(x_{n-1}))\}$ and $A_n=E_{f}(S_n)$ for any set sequence $S_n\in \X$.\label{local consistency}
\end{assumption}

In Assumption \ref{local consistency}, $\hat f(x;\D)$ represents the predictor at $x$ using data $\D$. This assumption stipulates that the true function value at a target point can be approximated with increasing precision by $\hat f$ constructed at evaluation points converging to this target, with the historically evaluated points and any additional data. 

\begin{assumption}[SNEB property of UQ]
The UQ $\hat\sigma(\cdot;\cdot):\X\times E_{f}(\X)\to\mathbb R$ satisfies:

\begin{enumerate}

\item For any $x\in\X$ and finite-cardinality set sequence $S_n\subset\X$, if $\inf_{n} \Delta(x, S_{n}) > 0$, then $\liminf_{n\to \infty}\hat\sigma(x; A_{n}) > 0$ where $A_n=E_{f}(S_n)$.

\item For any convergent sequence $x_n\in \X$, i.e., $x_n\to x'$ for some $x'\in \X$, we have $\hat\sigma(x_n;A_{n-1}\cup \D_{n-1})\to0$, where $\D_{n-1}=\{(x_1, f(x_1)),\ldots,(x_{n-1},f(x_{n-1}))\}$ and $A_n=E_{f}(S_n)$ for any finite-cardinality set sequence $S_n\subset \X$.
\end{enumerate}\label{GNEB}
\end{assumption}

Note that the SNEB property in Assumption \ref{GNEB} on the UQ is exactly the same as Assumption \ref{basic assumptions}. This highlights the key role of UQ as the driver of exploration and ultimately solution convergence via the EW framework. However, by incorporating the SP via the AF (discussed momentarily), we can induce exploitation to enhance practical performances.

\begin{assumption}[Improvement property of AF]\label{improvement}
The AF $g_n(\cdot,\cdot):\mathbb R\times\mathbb R_+\to\mathbb R$ satisfies the following (where $p_n$ and $q_n$ are any real sequences):

\begin{enumerate}
\item $\liminf_{n\to\infty} g_n(p_n,q_n)>0$ if $\liminf_{n\to\infty}p_n>-\infty$ and $\liminf_{n\to\infty}q_n > 0$.

\item $g_n(p_n,q_n)\to0$ if $\limsup_{n\to\infty}p_n \leq 0$ and $q_n\to0$.
\end{enumerate}
\end{assumption}

AF can be viewed as a channel to convert SP and UQ into EW. That is, the higher is the output of $g_n$ is, the more worthy to evaluate is the considered point. In Assumption \ref{improvement}, $p_n$ is the argument for the potential improvement regarding point estimation, and $q_n$ is the argument for the uncertainty. Part 2 of the assumption states that if there is, with eventual certainty, no improvement, then the worthiness to evaluate becomes zero. In contrast, part 1 stipulates that, as long as there is uncertainty, then there is some worthiness to evaluate the considered point (note that the condition $\liminf_{n\to\infty}p_n>-\infty$ there is largely a technicality that avoids $p_n$ being unboundedly negative). Finally, we allow AF to be step-dependent which is intended to make our PseudoBO framework general enough to cover common existing algorithms.

We are now ready to put together all the above ingredients into algorithmic consistency. First, for a set $\D\subset E_f(\X)$, denote $\Pi_{f}(\D)=\{y:(x,y)\in\D\text{\ for some\ }x\in\X\}$ as the projection of $\D$ onto the output dimension.

\begin{theorem}[From SP+UQ+AF to EW]
Suppose SP $\hat f$, UQ $\hat\sigma$, and AF $g_n$ satisfy Assumptions \ref{local consistency}, \ref{GNEB} and \ref{improvement} respectively. Suppose also that $f$ is continuous. Then the EW constructed by $W_n(x;\D_n)=g_n (\zeta(\hat f(x;\D_n)-\max \Pi_{f}(\D_n)), \hat\sigma(x;\D_n))$, where $\zeta(\cdot)$ is continuous and non-decreasing and $\zeta(0) \leq 0$, satisfies Assumption \ref{basic assumptions}.\label{PseudoBO main}
\end{theorem}

Based on Theorems \ref{main basic} and \ref{PseudoBO main}, we obtain the following guarantee:

\begin{corollary}[Algorithmic consistency via SP+UQ+AF]\label{cor: PseudoBO consistency via SP+UQ+AF}
Under the same assumptions as Theorem \ref{PseudoBO main}, for any compact $\X$, PseudoBO with EW constructed by $W_n(x;\D_n)=g_n(\zeta(\hat f(x;\D_n) - \max \Pi_{f}(\D_n))), \hat\sigma(x;\D_n))$ is algorithmically consistent.
\end{corollary}

\subsection{$(\delta,\epsilon)$-Relaxation of PseudoBO}
The above PseudoBO framework guarantees eventual popularization and subsequently algorithmic consistency. In this subsection, we relax the popularization requirement to only a certain resolution, i.e., any point in $\X$ has eventually evaluated points within say a $\delta$-sized neighborhood. This in turn leads to algorithmic consistency up to an error relating to $\delta$. We study this relaxation for two reasons. One is that, in many problems, it makes sense to aim for near-optimality instead of exact optimality, because a small optimality gap can play a negligible practical role which is not cost-effective to close in. Second, and more importantly, we will see that some SPs are in fact fundamentally accurate only up to a certain level of error. For such SPs, their pairing UQs are naturally SNEB up to an associated relaxation level, which in turn induces the $\delta$-relaxed eventual popularization.

We first state a relaxation of the SNEB property.

\begin{assumption}[$\delta$-relaxed SNEB property]
For a given $\delta\geq0$, $W_n(\cdot\ ;\ \cdot)$ satisfies the following:
\begin{enumerate}[leftmargin=*]

\item For any $x\in\X$ and finite-cardinality set sequence $S_n\subset\X$, if $\inf_{n} \Delta(x, S_{n}) > \delta$, then $\liminf_{n\to \infty}W_n(x; A_{n}) > 0$ where $A_n=E_{f}(S_n)$.

\item For any convergent sequence $x_n\in\X$, i.e., $x_n\to x'$ for some $x'\in\X$, we have $W_n(x_n;A_{n-1}\cup \D_{n-1})\to0$, where $\D_{n-1}=\{(x_1,f(x_1)),\ldots,(x_{n-1},f(x_{n-1}))\}$ and $A_n=E_{f}(S_n)$ for any finite-cardinality set sequence $S_n\subset\X$.
\end{enumerate}
\label{basic assumptions relaxed}
\end{assumption}

Compared to Assumption \ref{basic assumptions}, Assumption \ref{basic assumptions relaxed} only requires that $\liminf_{n\to \infty}W_n(x; A_{n}) > 0$ if $\inf_{n} \Delta(x, S_{n}) > \delta$ instead of $\inf_{n} \Delta(x, S_{n}) > 0$. That is, the point has positive EW if it is sufficiently, i.e., $\delta$ unit, far away from any eventually evaluated points. All other parts of the assumption remain the same as before. Assumption \ref{basic assumptions relaxed} is weaker than Assumption \ref{basic assumptions} when $\delta>0$, and reduces back to Assumption \ref{basic assumptions} when $\delta=0$.

With the above updated assumption, we obtain a corresponding relaxed version of algorithmic consistency. For this, we need to strengthen the continuity assumption of the objective function $f$ to Lipschitzness continuity. We call $f$ $L$-Lipschitz if $|f(x)-f(x')|\leq L\|x-x'\|$ for any $x,x'\in\X$ for some given constant $L>0$.

\begin{theorem}[$\delta$-relaxed algorithmic consistency of PseudoBO]
Suppose EW $W_n$ satisfies Assumption \ref{basic assumptions relaxed} and $\X$ is compact. Then:
\begin{itemize}
\item $\X$ is eventually populated by the PseudoBO iterates up to $\delta$-neighborhoods, i.e., for any $x\in\X$, we have $\inf_n\Delta(x,X_n)\leq\delta$ for $X_n=\{x_1,\ldots,x_n\}$ where $x_n$ is the $n$-th output iterate in the PseudoBO algorithm. 
\item Consequently, if furthermore $f$ is $L$-Lipschitz continuous, then PseudoBO is algorithmically consistent up to $L\delta$ error, i.e., the estimated optimal solution $\hat x_n^*\in\text{argmax}_{x\in X_n}f(x)$ satisfies $\liminf_nf(\hat x_n^*)\geq Z^*-L\delta$.
\end{itemize}\label{main basic relaxed}
\end{theorem}

Like Theorem \ref{main basic}, part 1 of Theorem \ref{main basic relaxed} concludes that the decision space $\X$ is eventually populated, but now up to $\delta$-sized neighborhoods. Part 2 then translates this popularization to algorithmic consistency, now with an error that depends on $\delta$ and the Lipschitz constant $L$.

Next, like Section \ref{sec:specialized}, we specialize our relaxed framework to the use of SP, UQ and AF. We first consider a relaxed version of local consistency of SP.

\begin{assumption}[$\epsilon$-relaxed local consistency of SP]
For a given $\epsilon\geq0$, the SP $\hat f(\cdot;\cdot): \X\times E_{f}(\X)\to\mathbb R$ satisfies that for any convergent sequence $x_n\in \X$, i.e., $x_n\to x'$ for some $x'\in \X$, we have 
$$\limsup_n|\hat f(x_n;A_{n-1}\cup \D_{n-1})-f(x')|\leq\epsilon$$
where $\D_{n-1}=\{(x_1,f(x_1)),\ldots,(x_{n-1},f(x_{n-1}))\}$ and $A_n=E_{f}(S_n)$ for any finite-cardinality set sequence $S_n\in \X$.\label{local consistency relaxed}
\end{assumption}

Compared to Assumption \ref{local consistency}, Assumption \ref{local consistency relaxed} stipulates that $\hat f$ is accurate only up to an $\epsilon$ error even when there is a sequence of evaluated points that gets infinitesimally close. That is, the predictor is fundamentally erroneous at level $\epsilon$. Note that, when $\epsilon=0$, Assumption \ref{local consistency relaxed} reduces back to Assumption \ref{local consistency}.

Next, the following is the relaxed version of SNEB for the UQ $\hat\sigma$, which is the same as that for EW in Assumption \ref{basic assumptions relaxed}.

\begin{assumption}[$\delta$-relaxed SNEB property of UQ]
For a given $\delta\geq0$, the UQ $\hat\sigma(\cdot;\cdot):\X\times E_{f}(\X)\to\mathbb R$ satisfies:
\begin{enumerate}[leftmargin=*]

\item For any $x\in\X$ and finite-cardinality set sequence $S_n\subset\X$, if $\inf_{n} \Delta(x, S_{n}) > \delta$, then $\liminf_{n\to \infty}\hat\sigma(x; A_{n}) > 0$ where $A_n=E_{f}(S_n)$.

\item For any convergent sequence $x_n\in\X$, i.e., $x_n\to x'$ for some $x'\in\X$, we have $\hat\sigma(x_n;A_{n-1}\cup \D_{n-1})\to0$, where $\D_{n-1}=\{(x_1,f(x_1)),\ldots,(x_{n-1},f(x_{n-1}))\}$ and $A_n=E_{f}(S_n)$ for any finite-cardinality set sequence $S_n\subset\X$.
\end{enumerate}
\label{GNEB relaxed}
\end{assumption}

Finally, the improvement property of AF remains the same under the relaxed framework as before, i.e., we would still use Assumption \ref{improvement} for AF. We then have the following guarantee.

\begin{theorem}[From SP+UQ+AF to EW under $(\epsilon,\delta)$-relaxation]
Suppose SP $\hat f$, UQ $\hat\sigma$, and AF $g_n$ satisfy Assumptions \ref{local consistency relaxed}, \ref{GNEB relaxed} and \ref{improvement} respectively. Suppose also that $f$ is continuous. Then the EW constructed by $W_n(x;\D_n)=g_n (\zeta(\hat f(x;\D_n)-\max \Pi_{f}(\D_n)-\epsilon), \hat\sigma(x;\D_n))$, where $\zeta(\cdot)$ is continuous and non-decreasing and $\zeta(0) \leq 0$, satisfies Assumption \ref{basic assumptions relaxed}.\label{PseudoBO main relaxed}
\end{theorem}

Compared to Theorem \ref{PseudoBO main}, in Theorem \ref{PseudoBO main relaxed} we use the relaxed versions of the local consistency and SNEB properties. Moreover, we use $\zeta(\hat f(x;\D_n)-\max \Pi_{f}(\D_n)-\epsilon)$ instead of $\zeta(\hat f(x;\D_n)-\max \Pi_{f}(\D_n))$ in the first argument of $g_n(\cdot,\cdot)$ when defining the EW. Recall that the first argument of $g_n(\cdot,\cdot)$ represents the potential improvement regarding point estimation. When the SP has a fundamental inaccuracy of $\epsilon$ unit, we naturally consider potential improvement only up to the same amount of prediction inaccuracy, thus leading to the extra $\epsilon$-reduction in this improvement calculation. 

Lastly, based on Theorems \ref{main basic relaxed} and \ref{PseudoBO main relaxed}, we obtain the following guarantee:

\begin{corollary}[Algorithmic consistency via SP+UQ+AF under $(\epsilon,\delta)$-relaxation]\label{cor: PseudoBO consistency via SP+UQ+AF relaxed}
Under the same assumptions as Theorem \ref{PseudoBO main relaxed}, for any compact $\X$ and assuming additionally that $f$ is $L$-Lipschitz, PseudoBO with EW constructed by $W_n(x;\D_n)=g_n(\zeta(\hat f(x;\D_n) - \max \Pi_{f}(\D_n))-\epsilon), \hat\sigma(x;\D_n))$ is algorithmically consistent up to $L\delta$ error.
\end{corollary}

Corollary \ref{cor: PseudoBO consistency via SP+UQ+AF relaxed} concludes that, when we define the potential improvement in a way that addresses the $\epsilon$ prediction inaccuracy, and the UQ is obtained with a $\delta$-relaxed SNEB property, the resulting PseudoBO procedure would have algorithmic consistency up to an $L\delta$ error. Note that when $\epsilon=\delta=0$, Corollary \ref{cor: PseudoBO consistency via SP+UQ+AF relaxed} reduces back to the non-relaxed case in Corollary \ref{cor: PseudoBO consistency via SP+UQ+AF}. In fact, we note that the requirement of $\epsilon$ and $\delta$ are separate, i.e., the corollary holds even if only one of $\epsilon$ and $\delta$ is non-zero. That is to say, we can opt to aim for popularization up to $\delta$-neighborhoods, and consequently $L\delta$-optimality, when the SP has no fundamental error. Similarly, we might have an $\epsilon$-inaccurate SP, but we choose an UQ that has a precise SNEB property instead of being $\delta$-relaxed. This is indeed possible; however, as we will see in the next section, some natural choices of UQs that couple with an $\epsilon$-relaxed locally consistent SP would only bear the $\delta$-relaxed SNEB property. This latter behavior deems the relaxation on both the local consistency and the SNEB property important and naturally coupled.

We close this section by explaining why it is important to consider SPs that are $\epsilon$-relaxed locally consistent. This arises from the fact that many common machine learning predictors contain hyperparameters typically needed to be tuned in relation to the sample size. Such tuning would ensure the predictor is locally consistent, but only if the design points are sampled according to certain distributions. As we define our local consistency (Assumptions \ref{local consistency} and \ref{local consistency relaxed}) in a way that is free of any distributional assumption on the $\{x_n\}$ sequence, standard tuning approaches that aim for consistency would not apply. Our $\epsilon$-relaxation serves to remedy this issue since, to achieve the relaxed local consistency, we can simply choose a fixed hyperparameter value instead of scaling it with the number of evaluated points. 

\section{The PseudoBO Cookbook}\label{sec:recipe}

\begin{figure} 
\centering
\includegraphics[width=\textwidth]{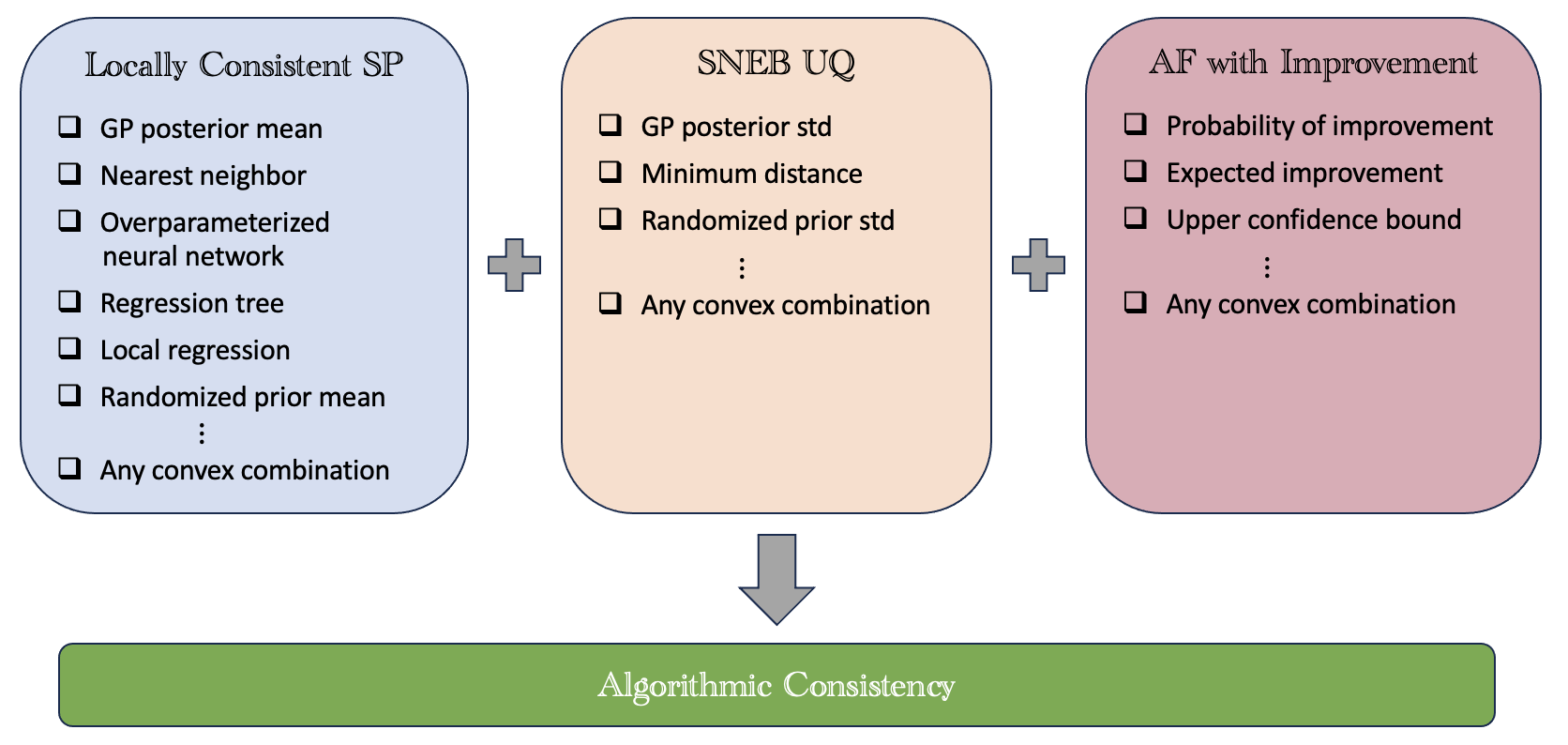}
\caption{\small{A general recipe for configuring a PseudoBO algorithm.}} 
\label{fig:PseudoBO}
\end{figure}

We present a range of examples for SP, UQ and AF to demonstrate the generality of PseudoBO and how it applies to existing algorithms as well as new ones. Importantly, it also paves the way for us to select practically superior algorithms. Our results can be summarized as a PseudoBO recipe in Figure \ref{fig:PseudoBO}.


\subsection{SP with Local Consistency}\label{sec:SP examples}

The essence of local consistency is that the SP can correctly estimate the objective value with sufficient data around, given continuity of the objective. An exemplifying type of locally consistent SP is the class of interpolating models, which give prediction values exactly as the evaluated values for all the evaluated points, and similar to the values of the surrounding evaluated points otherwise.  GP posterior mean, nearest neighbor and over-parameterized neural networks are examples. For our discussion below, we recall the notation $X_n=\{x_1,\ldots,x_n\}\subset\X$ as a finite set of points that we select to evaluate, and $\D_n=\{(x_1,f(x_1)),\ldots,(x_n,f(x_n))\}$ as the collection of evaluated pairs.


We start with GP posterior mean predictor. To define this, we first choose a positive semi-definite kernel function, $\mathcal{K}: \R^{d} \times \R^{d} \to \R$, and a mean function, $\mu_0: \R^{d} \to \R$. The mean function is commonly a constant function $\mu_0(x) = 0$ for centered GPs or a low-order polynomial (\cite{frazier2018tutorial}).
Given data $\D_n$, the GP posterior mean predictor is defined as 
\begin{equation}
\hat{f}_{\text{GP}}(x_0; \D_n) = \mathcal{K}(x_0, X_n)\mathcal{K}(X_n, X_n)^{-1}(f(X_n) - \mu_0(X_n)) + \mu_0(x_0),\label{GP SP}
\end{equation}
where $\mathcal{K}(x_0, X_n):= [\mathcal{K}(x_0, x_1), ..., \mathcal{K}(x_0, x_n)]$, $\mathcal{K}(X_n, X_n):= [\mathcal{K}(x_i, x_j)]_{1\leq i,j \leq n}$, $\mu_0(X_n):= [\mu_0(x_1), ..., \mu_0(x_n)]$ and $f(X_n):=[f(x_1), ..., f(x_n)]$, for a test point $x_0\in\X$. The kernel function $\mathcal K$ and mean function $\mu_0$ correspond to the prior covariance and mean of the GP, and with data $\D_n$ we output the posterior mean of the GP as the SP \eqref{GP SP}. Note that
the GP posterior mean predictor is an interpolator. Specifically, for any observed data point $(x_i, f(x_i)) \in \D_n$, we have $\hat{f}_{GP}(x_i; \D_n) = f(x_i)$. The intuition behind this is that $K(x_i, X_n)K(X_n, X_n)^{-1} = e_i^{\intercal}$, leading to $\hat{f}_{\text{GP}}(x_i; \D_n) = e_i^{\intercal} (f(X_n) - \mu_0(X_n)) + \mu_0(x_i) = f(x_i)$.

\begin{proposition}[Local consistency of GP mean predictor]\label{prop:GP local consistent} Assume a GP's covariance function $\mathcal{K}$ is continuous and positive definite. Also assume $f \in \mathcal{H}$, the RKHS induced by $\mathcal{K}$. Then the GP posterior mean is locally consistent.
\end{proposition}

Note that we do not make any probabilistic assumption regarding the black-box function in Proposition \ref{prop:GP local consistent}. This is the essence of our PseudoBO framework, that we only extract properties needed to guarantee algorithmic convergence instead of using probabilistic assumptions from the Bayesian framework.

Next, we consider the nearest neighbor predictor
\begin{equation}
\hat{f}_{\text{NN}}(x_0;\D_n) = f(\argmin_{x\in X_n} \norm{x - x_0})
\end{equation}
for a test point $x_0\in\X$.

\begin{proposition}[Local consistency of nearest neighbor]\label{prop:Nearest local consistent}
Assume $f$ is continuous. Then the nearest neighbor predictor is locally consistent.
\end{proposition}

Next, a neural network predictor at point $x_0$ is given by:
\begin{equation}
\hat{f}_{\text{Net}}(x_0; \D_n) = f_{\phi_N, \mathbf{H}_N^*} \circ f_{\phi_{N-1}, \mathbf{H}_{N-1}^*}\circ...\circ f_{\phi_1, \mathbf{H}_1^*} (x_0),
\end{equation}
where $f_{\phi_i, \mathbf{H}_i}(x'):= \phi_i({\mathbf{H}_i} [x', 1])$, $\mathbf{H}_1 \in \R^{d_{1} \times (d+1)}$, $\mathbf{H}_{i} \in \R^{d_{i+1} \times (d_{i}+1)}$ for $i \geq 2$, $N$ denotes the number of layers,  $\phi_i$  denotes an activation function,
and $(\mathbf{H}_1^*,\ldots,\mathbf{H}_N^*) := \argmin_{\mathbf{H}_1,\ldots,\mathbf{H}_N} \frac{1}{n}\sum_{k=1}^{n}(f_{\phi_N, \mathbf{H}_N} \circ f_{\phi_{N-1}, \mathbf{H}_{N-1}}\circ...\circ f_{\phi_1, \mathbf{H}_1} (x_k) - f(x_k))^2$. Moreover, by over-parametrization we mean that for any dataset $\D_n$, the hyperparameters, such as the depth of the network $N$ and the hidden dimensions $\{d_i\}_{1\leq i\leq N}$ can be adjusted so that $(\mathbf{H}_1^*,\ldots,\mathbf{H}_N^*)$ 
can be chosen to perfectly fit the data $\D_n$, i.e., $\hat{f}_{\text{Net}}(x; \D_n)=f(x)$ for any $x\in X_n$.

\begin{proposition}[Local consistency of over-parameterized neural network]\label{prop:ONN local consistent} Assume a neural network predictor is $L$-Lipschitz and over-parameterized. Then it is locally consistent.
\end{proposition}

In fact, a scrutiny at the proof of Proposition \ref{prop:ONN local consistent} reveals that its conclusion holds as long as the SP is $L$-Lipschitz and over-parameterized. That is, we do not need the neural network structure in our argument. Nonetheless, neural network offers a natural class of SP where the needed properties can be attained.


Another type of locally consistent SP is the family of models that, in some sense, average the values of the surrounding evaluated points, and the prediction values at any evaluated points are not necessarily the evaluated values. This includes, for instance, regression trees and local regression.


We start with the regression tree predictor $\hat{f}_{\text{Tree}}(x_0; \D_n)$, where the tree structure is parameterized by some splitting rule $\theta$. The tree predictor at point $x_0$ is given by:
\begin{equation}\label{eq: decision_tree}
\hat{f}_{\text{Tree}}(x_0; \D_n) = \sum_{i=1}^{n} w_i(x_0;\theta) f(x_i),
\end{equation}
and $R_{l(x_0, \theta)}$ denotes the region covered by the leaf $l$, determined by both the input $x_0$ and $\theta$. If $\sum_{i=1}^{n}\1\{x_i \in R_{l(x_0, \theta)}\}=0$, then we define 
$\hat{f}_{\text{Tree}}(x_0; \D_n) = \frac{1}{n}\sum_{i=1}^n f(x_i)$.



\begin{proposition}[$\epsilon$-relaxed local consistency of regression tree]\label{prop:Tree local consistent} 

Assuming $f$ is $L$-Lipschitz, the regression tree predictor (\ref{eq: decision_tree}) parameterized by splitting rule $\theta$ is $2L\cdot \max_{x\in \X}\text{diam}(R_{l(x, \theta)})$-relaxed locally consistent, where $\text{diam}(R)$ for a region $R$ refers to $\max_{x,y \in R}\|x-y\|.$
\end{proposition}

Proposition \ref{prop:Tree local consistent} stipulates that if we configure the regression tree such that all its leaves are small, then it is $\epsilon$-relaxed locally consistent with an $\epsilon$ that scales with this small leaf size and the Lipschitz constant of $f$.

Next we present the local regression predictor:
\begin{equation}
\hat f_{\text{LR}}(x_0;\D_n)=\frac{\sum_{i\leq n}K\left(\frac{\|x_0-x_i\|}{h}\right)f(x_i)}{\sum_{i\leq n}K\left(\frac{\|x_0-x_i\|}{h}\right)}\label{kernel def}
\end{equation}
where $K$ is a non-negative function with bounded support on $[0,M]$ for some $M>0$, i.e., $K(y)=0$ for $y>M$ and $K(y)>0$ for $y\leq M$. The hyperparameter $h>0$ is the bandwidth. Definition \eqref{kernel def} is well-defined as long as there is an $x_i$ for $i\leq n$ such that $\|x_0-x_i\|\leq Mh$. If the latter does not hold, we output a prefixed constant, e.g., 0, as the prediction value. Note that we can modify all the above slightly if the bounded support is on $[0,M)$, i.e., when $y=M$, $K(y)=0$ instead of being positive, and all the discussion can be easily adopted accordingly. Definition \ref{kernel def} mimics the Nadaraya-Watson estimator. As a simple example, we can take $K(y)$ to be a positive constant for $y\in[0,M]$. 


\begin{proposition}[$\epsilon$-relaxed local consistency of local regression]\label{prop:KR local consistent} For local regression predictor \eqref{kernel def}, where $K$ is a non-negative function with bounded support on $[0,M]$ for some $M>0$, and $h>0$ is the bandwidth. Assume $f$ is $L$-Lipschitz. Then the local regression estimator is $LMh$-relaxed locally consistent.
\end{proposition}

In typical statistical contexts, the hyperparameters $\theta$ and $h$ in regression trees and local regression scale with the sample size $n$. However, here we take $\theta$ and $h$ to be fixed. As discussed before, unlike in conventional statistical contexts, we make no assumption on how $x_i$ are distributed; in fact, $x_i$ can be selected according to a completely deterministic process in the PseudoBO algorithm, and thus the standard way of tuning these hyperparameters does not apply. Instead, we fix them and aim for $\epsilon$-relaxed local consistency, i.e., we pay a small price on the convergence due to fixing these hyperparameters, in exchange for our capability to bypass the tuning issue.

Local consistency can be preserved if we apply the locally consistent SP together with a ``randomized prior". This randomized prior means we first sample from a random field to obtain a function, then we perturb the function values of the evaluated points by this randomly generated function and construct a predictor. When we output the prediction values, we compensate the prediction from the perturbed data by this function. This idea is more relevant when we discuss UQ, where the randomized prior serves as a natural artifact to quantify uncertainty. Here, we want to point out that an average over these perturbed predictions also give rise to valid SPs.


More precisely, let $r:\X\to\mathbb R$ be a continuous function randomly generated from a generating distribution $\mathcal R$. Suppose a ``base" predictor is $\hat f$. The randomized prior mean prediction is
\begin{equation}
    \hat{f}_{\text{RP}}(x_0;\D_n) = \E_{r\sim \mathcal{R}} [\hat f^{(r)}(x_0;\D_n)], \label{RP mean}
\end{equation}
where $\hat f^{(r)}(x_0;\D_n)=r(x_0) + \hat{f}(x_0; \D_n^{(r)})$ is a ``perturb-then-compensate" predictor using the function $r$, and $\D_n^{(r)} := \{(x_i, f(x_i) - r(x_i))\}$ is the $r$-perturbed data set.

\begin{proposition}[Preservation of $\epsilon$-relaxed local consistency of randomized prior mean]\label{prop:RP local consistent} Suppose $r$ is a continuous function randomly generated from the generating distribution $\mathcal R$, and $\hat f^{(r)}$ is $\epsilon$-relaxed locally consistent in predicting $f$ for almost surely any $r$. Also assume $f,\hat f^{(r)}$ are uniformly bounded. Then the randomized prior mean \eqref{RP mean} is $\epsilon$-relaxed locally consistent.
\end{proposition}

Lastly, we have another useful preservation result for local consistency. Specifically, if we have a class of ($\epsilon$-relaxed) locally consistent SPs, then their convex combination, which we call a hybrid SP, will also enjoy ($\epsilon$-relaxed) local consistency. 

\begin{proposition}[Preservation of $\epsilon$-relaxed local consistency of hybrid SP]\label{prop:Hybrid local consistent} Given a finite set of $\epsilon$-relaxed locally consistent SPs $\hat{\mathcal{F}} := \{ \hat{f_1}, \hat{f_2}, ...\}$, the hybrid SP $\sum_{i = 1}^{|\hat{\mathcal{F}}|} \alpha_i\hat f_i$, where $\sum_{i = 1}^{|\hat{\mathcal{F}}|}\alpha_i = 1$ and $\alpha_i\geq0$, is $\epsilon$-relaxed locally consistent.
\end{proposition}

Clearly, Propositions \ref{prop:RP local consistent} and \ref{prop:Hybrid local consistent} apply to the basic notion of local consistency in Definition \ref{local consistency} as a special case when taking $\epsilon=0$. Moreover, in the latter case, Proposition \ref{prop:Hybrid local consistent} can be strengthened to allow $\alpha_i$ to be negative by using a similar proof.

\subsection{UQ with SNEB Property}

SNEB in a UQ entails that the quantified uncertainty at the queried points decreases to 0 as the surrounding evaluated points populate, while the uncertainty stays positive for any unexplored area. To verify SNEB for a UQ that is associated with a particular SP, it is typically the case that if the SP is locally consistent and that the UQ is a ``truthful" representation of the uncertainty, then part 2 of the SNEB property in Assumption \ref{GNEB} holds. On the other hand, the verification of part 1 of that assumption would require further conditions, as it is about the unexplored region where local consistency property does not touch upon.

We first consider the GP posterior variance (or standard deviation). This UQ, at point $x_0$, is given by:
\begin{equation}
\hat{\sigma}_{\text{GP}}^2(x_0;  \D_n):=\mathcal{K}(x_0, x_0) - \mathcal{K}(x_0, X_n){\mathcal{K}(X_n, X_n)^{-1}}\mathcal{K}(X_n, x_0)\label{eq: GP UQ}
\end{equation}
which is derived through modeling $f$ as a GP with prior covariance represented via $\mathcal K$ (\cite{sacks1989design}). Note that $\hat{\sigma}_{\text{GP}}^2(x_0;  \D_n)$ does not depend on the prior mean function $\mu_0$ or the labels in the data as seen from (\ref{eq: GP UQ}).


Furthermore, we call a covariance function stationary if it has the property $\K(x, y) = \K(x-y)$. That is, the function evaluations on the two inputs $x$ and $y$ only depend on their difference. This property is commonly imposed for modeling black-box functions, and popular examples are Gaussian covariance and Matérn covariance:
\begin{equation}
    \K_{Gaussian}(x,y) = \exp\{-\frac{\|x-y\|^2}{2h^2}\},
\end{equation}
where the bandwidth $h$ is a positive hyperparameter, and
\begin{equation}
    \K_{Matern}(x,y) = \frac{2^{1-\nu}}{\Gamma(\nu)}(\sqrt{2\nu}\frac{\|x-y\|}{\rho})^{\nu}K_{\nu}(\sqrt{2\nu}\frac{\|x-y\|}{\rho}),
\end{equation}
where $\Gamma$ is the gamma function, $K_{\nu}$ is the modified Bessel function of the second kind, and $\rho$, $\nu$ are positive hyperparameters for the covariance function.

Finally, the spectral density of a GP characterizes its frequency domain. For a stationary GP, the covariance function $\K(\tau)$, where $\tau:= x-y$, can be transformed into the frequency domain using the Fourier transform, which gives us the spectral density. Mathematically, the spectral density $\mathcal{S}(\omega)$ is defined as the Fourier transform of the covariance function:
$$\mathcal{S}(\omega) = \int_{-\infty}^{\infty}\K(\tau)\exp(-i\omega \tau)\text{d}\tau.$$

With all the above, we have the following guarantee:



\begin{proposition}[SNEB of GP posterior standard deviation]\label{prop:GP GNEB} Under the same assumption as in Proposition \ref{prop:GP local consistent}, and additionally, the GP is stationary and has spectral density $\mathcal{S}$, with the property that $\mathcal{S}^{-1}$ has at most polynomial growth, then the GP posterior standard deviation $\hat\sigma_{\text{GP}}(x_0;\D_n)$ is SNEB.
\end{proposition} 

Proposition \ref{prop:GP GNEB} largely follows from the seminal GP result in  \cite{vazquez2010convergence}. Next, we also have that the minimum distance UQ is SNEB.

\begin{proposition}[SNEB of minimum distance]\label{prop:MD GNEB}
$\Delta$ is SNEB.
\end{proposition}



Next we present a UQ utilizing randomized prior, which we have briefly discussed in the SP examples in Section \ref{sec:SP examples}. This method first randomly generates a continuous function $r$ from a generating distribution $\mathcal{R}$. We fit the SP $\hat{f}^{(r)}(x_0; \D_n) = r(x_0) + \hat f(x_0; \D^{(r)}_n)$, where $\hat f$ is a base SP and $\D^{(r)}_n$ is a data set perturbed by $r$, i.e., $\D_n^{(r)} := (x_i, f(x_i) - r(x_i))_{i=1}^{n} $. Upon repeating the sampling of $r$ many times, the UQ can be computed as $(\Var_{r\sim \mathcal{R}} [\hat{f}^{(r)}(x_0; \D_n)])^{\nicefrac{1}{2}}$.

Here is the rationale of this randomized prior UQ. Suppose a test point $x_0$ is surrounded by many evaluated points. Then, if $f$ is continuous and $\hat f$ is locally consistent, we have $\hat f(x_0; \D^{(r)}_n)$ roughly equal to $f(x_0)-r(x_0)$, so that $\hat{f}^{(r)}(x_0; \D_n)$ becomes roughly $f(x_0)$ by canceling out the $r(x_0)$. Thus, $\Var_{r\sim \mathcal{R}} [\hat{f}^{(r)}(x_0; \D_n)]$ becomes close to zero. This signals a low uncertainty at $x_0$. On the other hand, suppose $x_0$ has no surrounding evaluated points, or in other words the region surrounding $x_0$ is sparsely evaluated. Then, $\hat f(x_0; \D^{(r)}_n)$ would have weak reliance on $r$. For example, $\hat f$ could just output a constant for $x_0$ when there is no neighboring evaluated points. In this case, $\hat{f}^{(r)}(x_0; \D_n)$ will equal $r(x_0)$, which has a high variance. This signals a high uncertainty at $x_0$. This ``perturb-then-compensate" principle of randomized prior thus naturally distinguishes between densely and sparsely evaluated regions.
 
As far as we know, the concept of randomized prior appears initially in \cite{osband2018randomized} to quantify uncertainty in deep reinforcement. Compared to the bootstrap \cite{efron1982jackknife}, this approach appears procedurally similar but conceptually different. In particular, randomized prior uses the ``perturb-then-compensate" principle to signal high uncertainty for unexplored regions, while the bootstrap is designed to quantify the statistical errors arising from data randomness. The latter arguably works only when data are sufficiently abundant, since its underlying resampling can accurately resemble the sampling distribution only in such situations. Moreover, the uncertainty it captures comes from data randomness, while in the PseudoBO setting there is no data randomness and the uncertainty comes purely from the lack of knowledge on the objective function. Finally, we mention that \cite{osband2018randomized} proposes randomized prior with neural network; however, we generalize it here to broader regressors and can pair it with any other locally consistent SPs.

Finally, if the SP is $\epsilon$-relaxed locally consistent instead of exactly locally consistent, we can use
\begin{equation}
((\Var_{r\sim \mathcal{R}} [\hat{f}^{(r)}(x_0; \D_n)])^{\nicefrac{1}{2}}-\epsilon)_+\label{RP UQ relaxed}
\end{equation}
for any point $x_0$, where $(y)_+$ is the positive part function, i.e., $=y$ if $y>0$ and $=0$ otherwise. This definition allows for the $\epsilon$ prediction error, so that when the prediction uncertainty signified by $(\Var_{r\sim \mathcal{R}} [\hat{f}^{(r)}(x_0; \D_n)])^{\nicefrac{1}{2}}$ is less than $\epsilon$, we take the UQ as 0 in the PseudoBO procedure. Correspondingly, the UQ is positive only if the test point $x_0$ is sufficiently far away from all evaluated points, giving rise to a $\delta$-relaxed notion of SNEB.

\begin{proposition}[$\delta$-relaxed SNEB of randomized prior standard deviation]\label{prop:RP GNEB}
Under the same assumptions as Proposition \ref{prop:RP local consistent}, and that $(\Var_{r\sim \mathcal{R}} [\hat{f}^{(r)}(x_0; \D_n)])^{\nicefrac{1}{2}}>\epsilon$ whenever $\inf_n\Delta(x_0;\D_n)>\delta$ for any $x_0$ and $\D_n$, for some fixed $\delta>0$. Then the UQ \eqref{RP UQ relaxed} is $\delta$-relaxed SNEB.
\end{proposition}

We can apply randomized prior variance to local regression SP defined in \eqref{kernel def}, giving the following:
\begin{corollary}[$\delta$-relaxed SNEB of randomized prior standard deviation for local regression]\label{prop:RP GNEB kernel}
Under the same assumptions as Proposition \ref{prop:KR local consistent}, suppose $r$ is a $\tilde L$-Lipschitz and uniformly bounded function randomly generated from the generating distribution $\mathcal R$, with $(\Var_{r\sim\mathcal R}[r(x)])^{\nicefrac{1}{2}}>(L+\tilde L)Mh$ for any $x\in\X$. Also assume $f$ is bounded. Then the UQ $((\Var_{r\sim \mathcal{R}} [\hat{f}^{(r)}(x_0; \D_n)])^{\nicefrac{1}{2}}-(L+\tilde L)Mh)_+$ for any point $x_0$, with base SP \eqref{kernel def}, is $Mh$-relaxed SNEB.
\end{corollary}

Finally, like our discussion of SP, we also have the preservation of ($\delta$-relaxed) SNEB for a combination of UQs which we call hybrid UQ: 
\begin{proposition}[Preservation of $\delta$-relaxed SNEB of hybrid UQ]\label{prop: hybrid UQ GNEB}
Given a finite set of $\delta$-relaxed SNEB UQs $\hat{\Sigma} := \{ \hat{\sigma_1}, \hat{\sigma_2}, ...\}$, the hybrid UQ $\sum_{i = 1}^{|\hat{\Sigma}|} \alpha_i\hat{\sigma_i}$, where $\sum_{i = 1}^{|\hat{\Sigma}|}\alpha_i = 1$ and $\alpha_i \geq 0$, is $\delta$-relaxed SNEB.   
\end{proposition}

Similar to our discussion of SP in Section \ref{sec:SP examples}, Proposition \ref{prop: hybrid UQ GNEB} reduces to the case of exact SNEB when $\delta=0$. On the other hand, unlike there, we can readily see from the proof of Proposition \ref{prop: hybrid UQ GNEB} that we can relax the convex combination of UQs to any linear combination of UQs with weights $\alpha_i \geq 0$ for all $i$ with at least one of the weights being strictly positive. Nonetheless, from a practical perspective, scaling the UQs by a multiplicative factor does not change the PseudoBO performance as long as we suitably rescale the AF function.

\subsection{AF with Improvement Property}
\vspace{-1mm}
The improvement property of AF, which signifies a zero EW for points that certainly lead to no improvement while positive for uncertain points, is satisfied by classical criteria such as probability of improvement (PI) and expected improvement (EI). These criteria can be written as
\begin{equation}
    g_n^{\text{PI}}(p_n , q_n) = \begin{cases}
                \Phi\left( \frac{p_n - \tau}{q_n}\right), & \text{if}\ q_n > 0,\\
                \1\{p_n - \tau  > 0\}, & \text{if}\ q_n = 0. \\
              \end{cases}\label{PI def}
\end{equation}
for PI, and 
\begin{equation}
    g_n^{\text{EI}}(p_n, q_n) = \begin{cases}
                q_n\phi(\frac{p_n - \tau}{q_n}) + (p_n - \tau) \Phi(\frac{p_n - \tau}{q_n}), & \text{if}\ q_n > 0,\\
                \max\{p_n - \tau, 0\}, & \text{if}\ q_n = 0. \\
              \end{cases}\label{EI def}
\end{equation}
for EI, where $\phi(\cdot)$ and $\Phi(\cdot)$ denote the standard normal density and distribution functions, and $\tau\geq0$ is a small hyperparameter to ensure smoothness at the ``boundary" between $q_n=0$ and $q_n>0$, especially in the PI case where we will set $\tau>0$. Typically, the potential improvement $p_n$ is defined as $\hat f(x;\D_n)-\max \Pi_{f}(\D_n)$, which is the value of the SP $\hat f$ above the best evaluated objective value $\max \Pi_{f}(\D_n)$, and the uncertainty estimate $q_n$ is taken as our UQ $\hat\sigma(x;\D_n)$. With these, \eqref{PI def} and \eqref{EI def} give
\begin{equation}\label{eq:PI definition}
    \text{PI}(x; \D_n) = \begin{cases}
                \Phi\left( \frac{\hat f(x;\D_n)-\max \Pi_{f}(\D_n) - \tau}{\hat{\sigma}(x;\D_n)}\right), & \text{if}\ \hat{\sigma}(x;\D_n) > 0,\\
                \1\{\hat f(x;\D_n)-\max \Pi_{f}(\D_n) - \tau  > 0\}, & \text{if}\ \hat{\sigma}(x;\D_n) = 0. \\
              \end{cases}
\end{equation}
and
\begin{align}\label{eq:EI definition}
    &\text{EI}(x; \D_n) \notag\\
    &=\begin{cases}
                \hat{\sigma}(x;\D_n)\phi(\frac{\hat f(x;\D_n)-\max \Pi_{f}(\D_n) - \tau}{\hat{\sigma}(x;\D_n)}) + (\hat f(x;\D_n)-\max \Pi_{f}(\D_n) - \tau) \Phi(\frac{\hat f(x;\D_n)-\max \Pi_{f}(\D_n) - \tau}{\hat{\sigma}(x;\D_n)}), \\
                \ \ \ \ \ \ \ \ \ \ \ \ \ \ \ \ \ \ \ \ \ \ \ \ \text{if}\ \hat{\sigma}(x;\D_n) > 0,\\
                \max\{\hat f(x;\D_n)-\max \Pi_{f}(\D_n) - \tau, 0\}, \\
                \ \ \ \ \ \ \ \ \ \ \ \ \ \ \ \ \ \ \ \ \ \ \ \ \text{if}\ \hat{\sigma}(x;\D_n) = 0. \\
              \end{cases}
\end{align}
The above definitions are precisely
$P(Z>\tau)$ and $E[(Z-\tau)_+]$ respectively, where $Z$ denotes a normal variable with mean $p_n=\hat f(x;\D_n)-\max \Pi_{f}(\D_n)$ and standard deviation $q_n=\hat\sigma(x;\D_n)$. To facilitate discussion, let $\tau=0$ for now. In the BO literature, $\hat f(x;\D_n)$ corresponds to the posterior mean and $\hat\sigma(x;\D_n)$ the posterior standard deviation, so that $P(Z>0)$ is the posterior probability under the GP model that a point $x$ has a higher objective value than $\max \Pi_{f}(\D_n)$, and $E[Z_+]$ is the expectation of this excess objective value. The following verifies that these classical PI and EI criteria satisfy our improvement property:





\begin{proposition}[Improvement Property of PI]\label{prop: PI improve}
With $\tau > 0$, PI defined as (\ref{PI def}) has the improvement property.
\end{proposition}

\begin{proposition}[Improvement Property of EI]\label{prop: EI improve}
With $\tau \geq 0$, EI defined as (\ref{EI def}) has the improvement property.
\end{proposition}

Another important acquisition approach is the upper confidence bound (UCB) widely used in online learning, defined as $\text{UCB}(x;\D_n) = \hat{f}(x;\D_n) + \beta_n \hat{\sigma}(x;\D_n)$ where $\beta_n$ is a step-dependent positive hyperparameter to trade-off exploitation and exploration. To connect to our AF notion, consider
\begin{equation}\label{eq:UCB definition}
    g_n^{\text{UCB}}(p_n,q_n) = \frac{p_n - \tau}{\beta_n} +  q_n
\end{equation}
where $\tau\geq0$ is again a hyperparameter like PI and EI described earlier (in fact, we can merely set $\tau=0$ in this case). Plugging in $p_n=\hat{f}(x;\D_n) - \max{\Pi_f(\D_n)}$ and $q_n=\hat\sigma(x;\D_n)$, we can see that \eqref{eq:UCB definition} is a rescaling of the more familiar form $\text{UCB}(x;\D_n)$ and thus the maximizer is retained. More precisely, we have $\argmax_{x\in \X} \hat{f}(x;\D_n) + \beta_n \hat{\sigma}(x;\D_n) = \argmax_{x\in \X} (\hat{f}(x;\D_n) - \max{\Pi_f(\D_n)} - \tau) + \beta_n \hat{\sigma}(x;\D_n) = \argmax_{x\in \X} \frac{p_n- \tau}{\beta_n} + q_n$ where $p_n=\hat{f}(x;\D_n) - \max{\Pi_f(\D_n)}$ and $q_n=\hat\sigma(x;\D_n)$.


\begin{proposition}[Improvement Property of UCB]\label{prop: UCB+ improve}
Suppose $\tau$ is fixed, $\beta_n \geq 0$ is a sequence that goes to $\infty$ as $n\to\infty$, and $p_n$ is bounded. Then UCB defined as (\ref{eq:UCB definition}) has the improvement property.
\end{proposition}

Note that we have assumed the sequence $p_n$ is bounded in Proposition \ref{prop: UCB+ improve}. When the objective function $f$ and SP $\hat f$ are bounded, then $p_n$ when plugged in as $\hat{f}(x;\D_n) - \max{\Pi_f(\D_n)}$ would be bounded as well, so that this additional assumption is readily achieved.

Finally, like SP and UQ, the improvement property is also preserved if we consider a combination of AFs, which we call hybrid AF:

\begin{proposition}[Improvement property of hybrid AF]\label{prop: hybrid improve} 
Given a finite set of AFs $\mathcal{G} := \{ g_1, g_2, ...\}$ with the improvement property, the hybrid AF $\sum_{i = 1}^{|\mathcal{G}|} \alpha_i g_i$, where $\sum_{i = 1}^{|\mathcal{G}|}\alpha_i = 1$ and $\alpha_i \geq 0$, has the improvement property.   
\end{proposition}

Similar to hybrid UQ, we can strengthen Proposition \ref{prop: hybrid improve} to relax the convex combination to any linear combination of AFs with weights $\alpha_i\geq0$ for any $i$ where at least one of these weights is strictly positive. Nonetheless, since a rescaling does not affect the maximizer of the AF, considering only convex combination is without loss of generality.

\section{From Theory to Implementation}\label{sec:implementation}

Our next goal is to leverage the PseudoBO recipe to construct empirically superior algorithms. We consider the following in choosing our ingredients for implementation:

\noindent\textbf{Measuring UQ Quality}.  To look for a good UQ, we propose a criterion called \textit{calibrated coverage rate (CCR)}, which considers the quality of both SP and UQ. Given a training set $\D_{\text{train}}:= (X_{\text{train}}, Y_{\text{train}})$, a validation set $\D_{\text{val}}:= (X_{\text{val}}, Y_{\text{val}})$, and a test set $\D_{\text{test}}:= (X_{\text{test}}, Y_{\text{test}})$, suppose we have a pretrained SP $\hat{f}(\cdot; \D_{\text{train}})$ and UQ $\hat{\sigma}(\cdot; \D_{\text{train}})$. We compute the test-set coverage rate of a ``prediction interval" $[\hat{f}(x; \D_{\text{train}}) - \lambda_{\text{val}}\hat{\sigma}(x; \D_{\text{train}}), \hat{f}(x; \D_{\text{train}}) + \lambda_{\text{val}}\hat{\sigma}(x; \D_{\text{train}})]$ by
\vspace{-1mm}
\begin{equation*}
    \pr_{(x, y)\sim\D_{\text{test}}}(x \in [\hat{f}(x; \D_{\text{train}}) - \lambda_{\text{val}}\hat{\sigma}(x; \D_{\text{train}}), \hat{f}(x; \D_{\text{train}}) + \lambda_{\text{val}}\hat{\sigma}(x; \D_{\text{train}})]),
\end{equation*}
where $\lambda_{\text{val}} = \min_{\lambda \geq 0} \lambda$ is chosen such that $\pr_{(x, y)\sim\D_{\text{val}}}(x \in [\hat{f}(x; \D_{\text{train}}) - \lambda\hat{\sigma}(x; \D_{\text{train}}), \hat{f}(x; \D_{\text{train}}) + \lambda\hat{\sigma}(x; \D_{\text{train}})]) = 1$, which ensures that the prediction interval is just wide enough to perfectly cover the validation set. This can be done by a bisection algorithm, with its details in Appendix \ref{appendix: CCR_details}. The higher the coverage is on the test set, the better is the SP+UQ intuited. If two SP+UQ combinations share the same coverage, we desire the one with narrower prediction intervals, as it indicates less conservativeness.


We consider four different SP+UQ pairs (Table 1) and compare their qualities via CRRs.

\begin{table}[htbp!]
\centering
\scalebox{0.8}{
\begin{tabular}{ccc}
\toprule
\textbf{Method Name} & \textbf{SP} & \textbf{UQ} \\ \hline
GP & GP posterior mean & GP posterior std \\ 
NN + MD & Nearest neighbor & Minimum distance \\ 
RP & Randomized prior mean & Randomized prior std \\ 
LR + Hyb & Local regression & Hybrid \\ 
\bottomrule
\end{tabular}
}
\caption{All combinations of SP + UQ. The randomized prior method here has base SP as local regression. Hybrid UQ is defined as a convex combination of minimum distance and randomized prior std.}
\label{tab:methods_comparison}
\end{table}
We test on three one-dimensional (1D) benchmark functions: 
\begin{enumerate}
    \item $f_1(x) = (\sin(\pi w))^2 + (w-1)^2 \cdot (1 + \sin(2\pi w)^2)$, where $w := 1 + (x-1)/4$, $x\in [-10, 10]$
    \item $f_2(x) = -20e^{-0.2|x|} - e^{\cos(2\pi x)} + 20 - e$, $x\in [-10, 5]$
    \item $f_3(x) =\sin(10\pi x)/(2x) + (x-1)^4$, $x\in [0.5, 2.5]$
\end{enumerate}
We generate the training set $\D_{\text{train}}$, the validation set $\D_{\text{val}}$ and the test set $\D_{\text{train}}$ by uniformly sampling $15$, $15$ and $50$ points from the decision space at random, with their labels are evaluated by the objective function. The bisection method is used to find $\lambda_{\text{val}}$.

\begin{figure}[htbp!] 
\centering
\includegraphics[width=\textwidth]{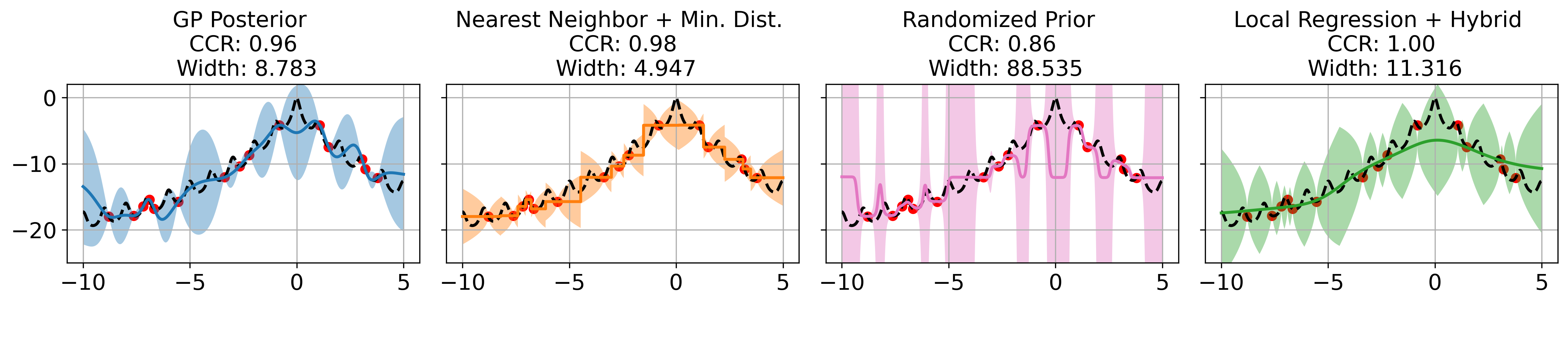}  

\caption{\small{A sample run of GP, NN + MD, RP, LR + Hyb to model the SP (the solid line) and the associated UQ (the shaded area). The training data points are marked with red dots.} }
\label{fig:coverage3}
\end{figure}

\begin{table}[htbp!] 
  \centering
  \scalebox{0.8}{
  \begin{tabular}{c|cc|cc|cc}
    \toprule
     & \multicolumn{2}{c|}{$f_1$} & \multicolumn{2}{c|}{$f_2$}  & \multicolumn{2}{c}{$f_3$}\\
    SP+UQ pair & CCR ($\uparrow$) & width ($\downarrow$) & CCR ($\uparrow$) & width ($\downarrow$) & CCR ($\uparrow$) & width ($\downarrow$) \\
    \midrule
    GP & $\underline{0.948\ (0.075)}$ & $7.91\ (2.48)$ & $\underline{0.964\ (0.023)}$ & $8.35\ (1.86)$ &  $0.936\ ( 0.041)$ & $2.33\ (0.87)$ \\
    \midrule
    NN + MD & $0.912\ (0.098)$ & $7.61\ (2.99)$ & $0.956\ (0.041$) & $5.884\ (0.792)$ & $\underline{0.944\ (0.03)}$ & $1.59\ (0.04)$\\
    \midrule
    RP & $0.900\ (0.081)$ & $9970.21\ (12026.52)$ & $0.900\ (0.077)$ & $7258.27\ (14274.74)$ & $0.708\ ( 0.105)$ & $6900.54\ (2553.29)$\\
    \midrule
    LR + Hyb & $\mathbf{0.968\ ( 0.020)}$ & $8.34\ (2.29)$  & $\mathbf{0.976\ (0.032)}$ & $15.39\ (5.39)$ & $\mathbf{0.960\ (0.052)}$ & $3.31 (1.02)$\\
    \bottomrule
  \end{tabular}}
  \vspace{1mm}
  \caption{\small{Coverage rate and width of calibrated ``prediction interval" on the test set. Results are repeated for 5 runs. The reported numbers are the empirical averages of these runs, with the values in brackets representing the empirical standard deviations. Bold numbers represent the best and the underlined numbers represent the runner-up.}}\label{tab: coverage}
\end{table}


From Table \ref{tab: coverage}, we see that LR + Hyb has the highest CCR across all three test cases. GP ranks as the runner-up in two cases, while NN + MD secures the runner-up position in one. In contrast, RP exhibits lower CCRs and wider calibrated intervals, possibly because while RP is designed to flag large uncertainty for unexplored areas, without bootstrapping on the training set, its uncertainty quantification ability for explored areas can be less precise. To be concrete, Figure \ref{fig:coverage3} shows some exemplified results on the three benchmark problems.

\noindent\textbf{Model Configuration}. We configure the following models: PseudoBO with SP and UQ as randomized prior mean and standard deviation (PseudoBO - RP) with the base SP as local regression, and PseudoBO with SP and UQ as local regression and a hybrid of minimum distance and randomized prior standard deviation (PseudoBO - LR + Hyb), defined as: 
\begin{equation}\label{eq:hybrid}
    \sigma_{\text{Hyb}}(x;\D_n) = \alpha \Delta(x, \D_n) + (1 - \alpha) \sigma_{\text{RP}}(x, \D_n),\hspace{1mm}
\end{equation}

The intuition of constructing this UQ stems from our previous discussion on measuring UQ quality. We observe that relying solely on either the minimum distance or the randomized prior standard deviation fails to achieve CCR performance comparable to that of GP.  Minimum distance considers only the nearest explored point, leaving out the effect of other nearby explored points. Specifically, as shown in Figure \ref{fig:coverage3}, GP posterior assigns higher uncertainty to the boundaries of the decision space —-where points are not surrounded by explored data —- than to the region around $x=0$, which is well-surrounded by explored points.  In contrast, the minimum distance metric assigns similar uncertainty across all three regions. On the other hand, the randomized prior standard deviation underestimates uncertainty in explored areas while overestimating it in unexplored regions. 

Theoretically, this hybrid UQ also satisfies the SNEB property according to Proposition \ref{prop: hybrid UQ GNEB}.
\\


\noindent\textbf{Optimizing EW.} From our empirical investigation, we recommend using a scrambled Sobol sequence and pick the point with the largest EW among the sequence. 

\section{Empirical Evaluations}\label{sec:experiments}

In this section, we perform extensive empirical experiments on a wide range of tasks: a toy example for testing the quality of UQ, $4$ synthetic black-box function optimization problems, $4$ hyperparameter tuning tasks, and $2$ robotic tasks. All these tasks are challenging and contain many local minima. The baseline models we consider in all the tasks comprise the standard BO, Random Search (RS) \cite{bergstra2012random}, SMAC \cite{hutter2011sequential}, TPE \cite{bergstra2011algorithms}, BORE \cite{tiao2021bore}, and LFBO \cite{song2022general}. In particular, BO is implemented using BoTorch \cite{balandat2020botorch}, and SMAC is implemented using AutoML \cite{JMLR:v23:21-0888}. 

For AF, the benchmarks BO, SMAC, and our PseudoBO variants use EI; TPE, BORE are designed to model PI; LFBO models EI. Additional details regarding our experiments can be found in Appendix \ref{appendix: exp details}. 


\subsection{Synthetic Black-Box Function Optimization}

\begin{figure}[htbp!] 
\centering
\includegraphics[width=.98\textwidth]{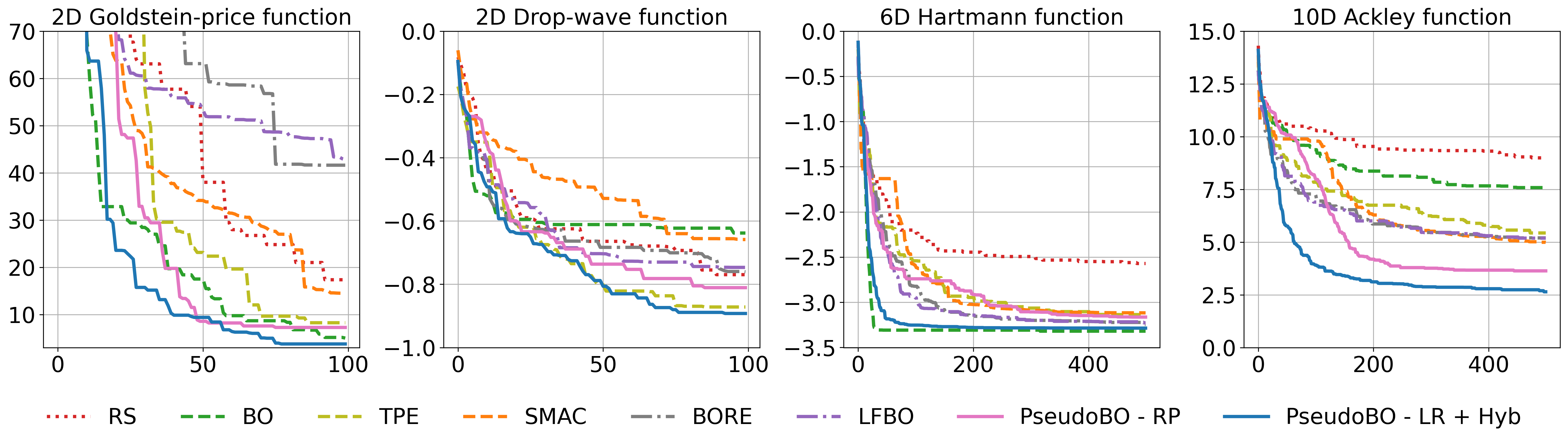} 
\caption{\small{Best objective queried against number of iterations for the synthetic black-box function minimization tasks. 
Each curve is an average over $10$ runs. 
}} 
\label{fig:synthetic}
\end{figure}

\begin{figure}[htbp!] 
\centering
\includegraphics[width=.98\textwidth]{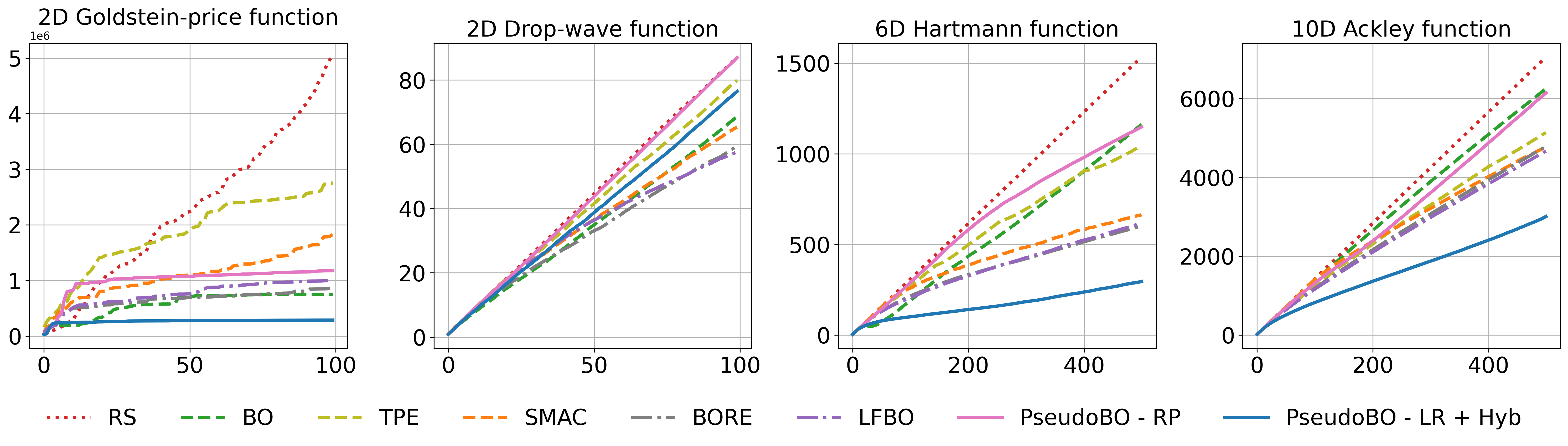}  
\caption{\small{
Cumulative regret in the synthetic black-box function minimization tasks. Each curve is an average over $10$ runs. 
}} 
\label{fig:synthetic_cumregt}
\end{figure}

We test our considered methods on the minimization of $4$ well-known benchmark functions, including a 2D Goldstein-price function, a 2D Drop-wave function, a 6D Hartmann function, and a 10D Ackley function. For all the methods, we run $100$ iterations with $5$ initial samples for the two $2$D functions, $500$ iterations with $10$ initial samples for the other two with higher dimensions. From Figure \ref{fig:synthetic}, we see that PseudoBO - LR + Hyb achieves the lowest objective in three out of four tasks within the query budget. Moreover, we report the cumulative regret in Figure \ref{fig:synthetic_cumregt}. PseudoBO - LR + Hyb is the best in three out of four tasks.


\begin{figure}[htbp!] 
\centering
\includegraphics[width=.98\textwidth]{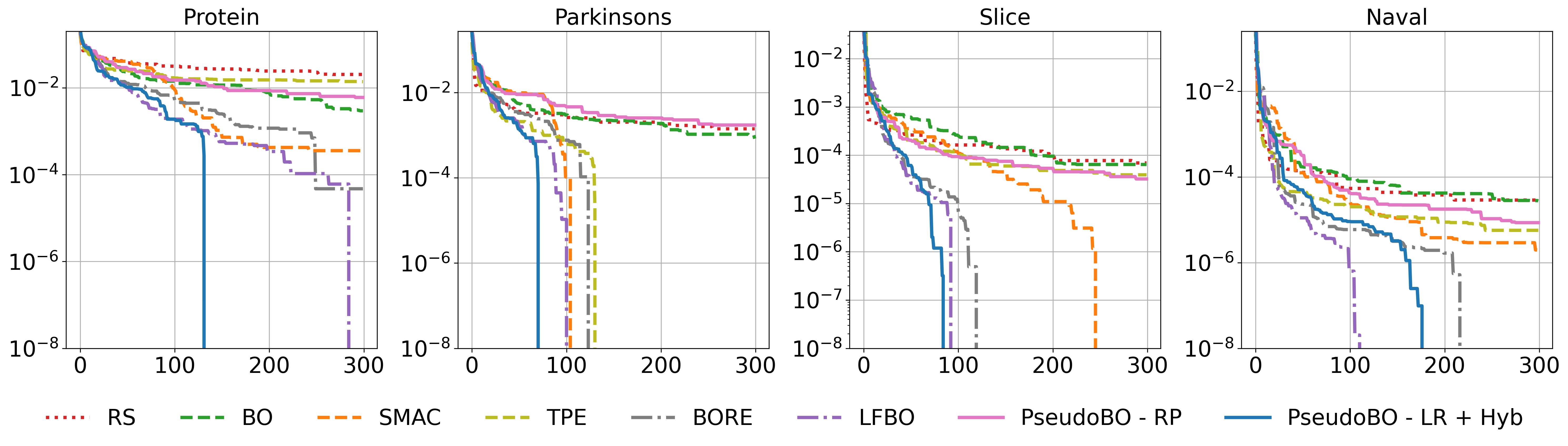}  
\caption{\small{Instant regret against number of iterations for the neural network tuning task. At each iteration, the instant regret is defined as the current best validation loss subtracted by the  validation loss of the optimal structure. Each curve is an average over $10$ runs. 
} }
\label{fig:nntuning}
\end{figure}

\subsection{Hyperparameter Tuning}

\begin{figure}[htbp!] 
\centering
\includegraphics[width=.98\textwidth]{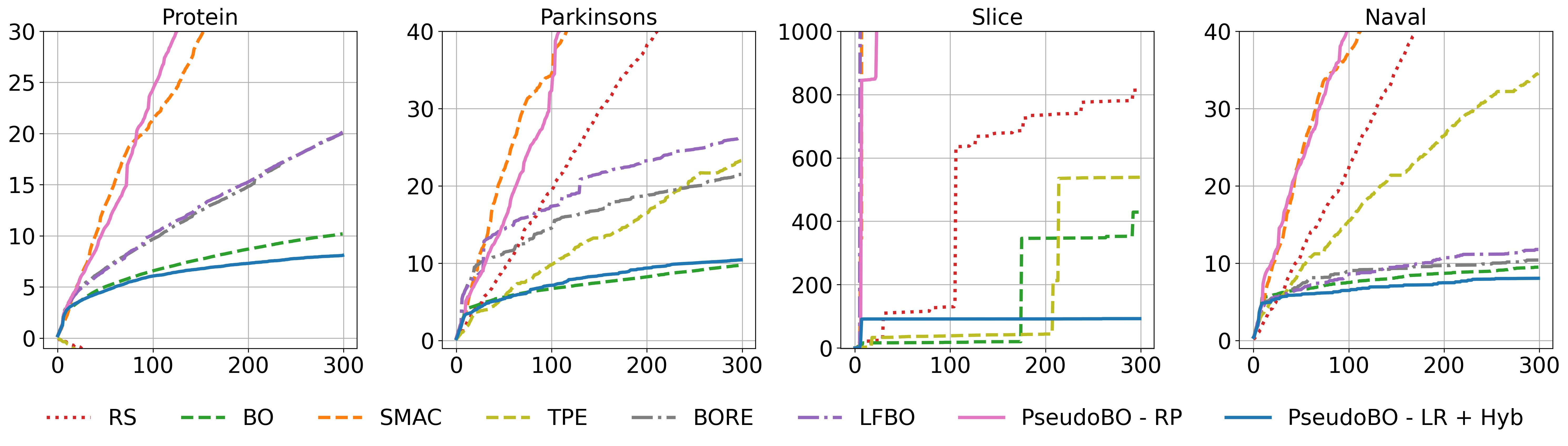}  
\caption{\small{Cumulative regret against number of iterations in the neural network tuning task. Each curve is an average over $10$ runs. 
} }
\label{fig:nntuning_cumregt}
\end{figure}

We test the methods on a more realistic task of neural network tuning \cite{klein2019tabular}. Our main goal here is to find the optimal set of hyperparameters for a two-layered fully connected neural network, to achieve the best training outcomes on four UCI datasets \cite{Dua:2017}.
There are $9$ hyperparameters, including initial learning rate, learning rate schedule, batch size, dropout rates, number of units, and activation function type for each of the two layers. This results in an expansive search space comprising $62,208$ possible configurations in total. 
Figure \ref{fig:nntuning} reveals that among all the methods, PseudoBO - LR + Hyb is the fastest one to uncover the optimal configuration in three out of four tasks. PseudoBO - RP underperforms in this task possibly because the randomized prior std underestimates the uncertainty in or near explored regions, thus neglecting good hyperparameters nearby. Relatively competitive models in this task are LFBO, BORE, and SMAC: LFBO achieves the best in the Naval task. BORE and SMAC are able to locate the optimal configurations in three out of four tasks. Besides, we also record the cumulative regret for all methods in Figure \ref{fig:nntuning_cumregt}. The plot shows that PseudoBO - LR + Hyb has sublinear cumulative regret in the iteration number and has the lowest cumulative regret in three out of four tasks. In the second task, BO has the lowest cumulative regret, and remains competitive across the other tasks.

\subsection{Robot Arm Pushing}\label{subsec: robotic arm}


This problem aims to control two robot arms with $14$ parameters to push two objects to target positions. The reward is evaluated by the ending positions of the two objects pushed by the two robot arms. See \cite{wang2018batched} for more details. In this task, we follow the procedure and setup in \cite{eriksson2019scalable}, conducting a batch of 50 queries in each iteration and performing optimization with a budget of $10000$ queries. All methods are initialized with $100$ queries. As demonstrated in Figure \ref{fig:rl} (1), all PseudoBO variants, particularly PseudoBO - LR + Hyb, converge significantly faster than other methods iteration-wise and eventually achieve the highest rewards. When measured in runtime, Figure \ref{fig:rl} (2) illustrates that PseudoBO - LR + Hyb takes merely 10 seconds to achieve a reward of around 9.75 followed by PseudoBO - RP and TPE, both achieveing reward around 8.5. Another point to note is that Hyperopt (the package implementing TPE) does not support batch evaluation. Consequently, TPE queries and proposes the next point sequentially, leveraging more information per query than other methods. This suggests that its performance might decline if batch evaluation were possible.

\subsection{Rover Trajectory Planning}\label{subsec: trajectory}

\begin{figure}[htbp!] 
\centering
\includegraphics[width=\textwidth]{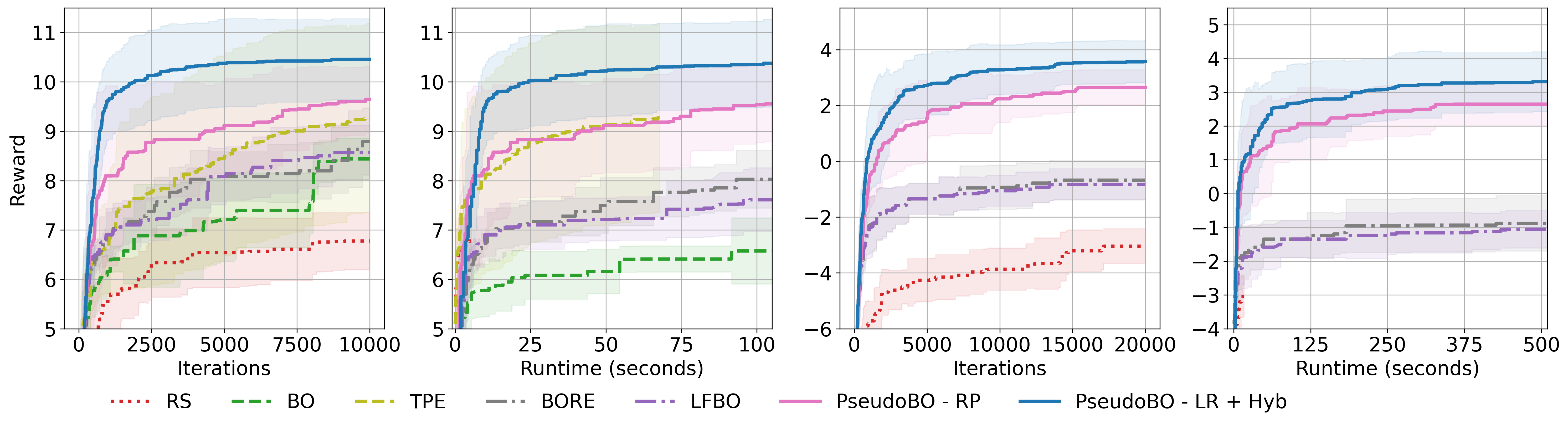}  
\caption{\small{Robot pushing task (first two) and Rover Trajectory task (last two). At each iteration, the best reward so far is recorded. Each curve is an average over $10$ runs. Shaded area is formed by $\pm$ one standard deviation. BO ceases at the 9,250-th query due to GPU being out of memory in the robot pushing task.}} 
\label{fig:rl}
\end{figure}

An additional robotic task we test is rover trajectory optimization by determining the locations of $30$ points on a 2D plane, where the final reward is estimated by the ending position of the rover and cost incurred by collision. See \cite{wang2018batched} for more details. We follow the procedure and setup in \cite{eriksson2019scalable} to perform a batch of queries of size 100 in each iteration and execute optimization within a $20,000$-query budget. All methods start with $200$ queries. Due to the memory issue, BO is difficult to run and not tested here. Besides, TPE is also excluded from the test, as sequentially querying and proposing next evaluations took excessively long time to finish $20,000$ queries. As illustrated in Figure \ref{fig:rl}(3), all PseudoBO variants, in terms of iterations, converge faster than the other methods and achieve the highest final rewards. A similar trend is observed when measured by runtime, as shown in Figure \ref{fig:rl}(4).


\subsection{Runtime Record}

We close our discussion by comparing the runtimes of the considered methods. Table \ref{tab: runtimes} enumerates the runtimes of all tested methods to complete the target number of queries in various tasks. Some entries are vacant because the corresponding methods were not included in our experiments due to memory or excessive runtime issues. In the larger-scale tasks, BO is eliminated due to memory exhausion, while SMAC and TPE are not included due to their packages' inability to support batch selection of candidates. 

In small-sized tasks, our record reveals that RS and TPE demonstrate shorter runtimes than others. Immediately following these are the methods including BORE, LFBO, PseudoBO - RP, and PseudoBO - LR + Hyb, which all exhibit comparable runtimes. Despite being not as fast as the two methods above, their performances are usually more promising in comparison (especially in the neural network tuning task). Finally, both BO and SMAC consistently take the longest to complete the designated number of queries.

\begin{figure}[htbp!]  
\centering
\includegraphics[width=\textwidth]{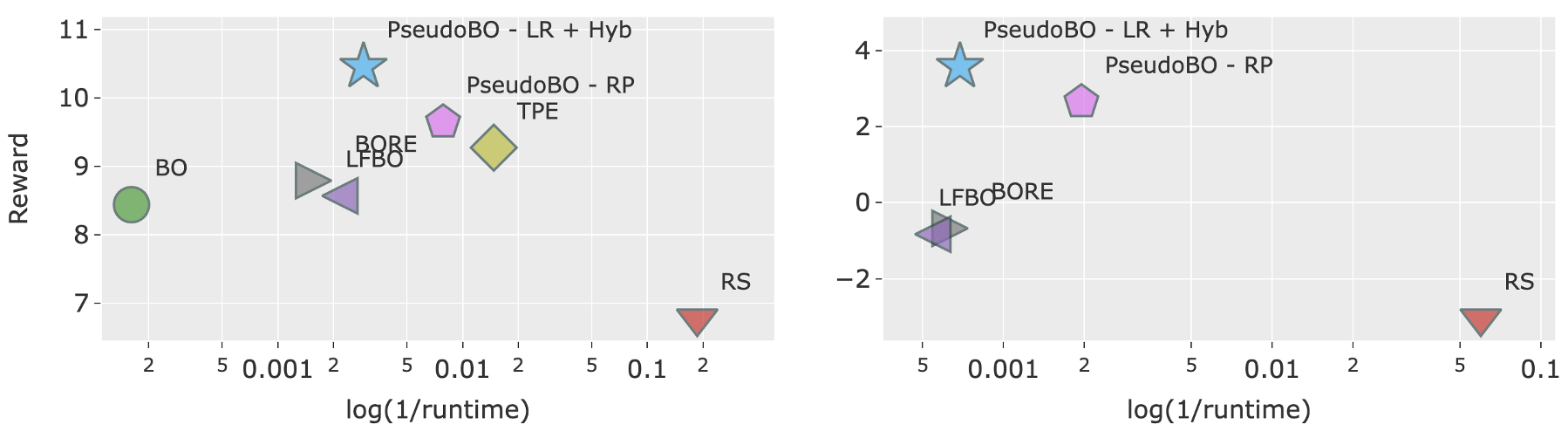}  
\caption{\small{Pareto plots of runtimes vs best objective values attained eventually in the robotic tasks (left for robot pushing and right for rover trajectory). In these visual representations, the effectiveness of a method increases alongside its position towards the upper right quadrant. The x-axis denotes the runtime with the transformation of $\log(\frac{1}{\text{runtime}})$ for the purpose of visualization.} }
\label{fig:pareto}
\end{figure}

As the query budget grows up to tens of thousands in the robotic tasks, RS and TPE still stand in the fastest tier in terms of the runtime, immediately followed by the PseudoBO family. Notably, the scalability of the latter becomes even more apparent in these tasks in the sense that the time gap with TPE narrows and the PseudoBO family outperforms in terms of attained objective value significantly. Compared with other methods, in the 14D robotic task, the fastest PseudoBO model (PseudoBO - RP) is 4X faster than BORE, 4X than LFBO, and \emph{48X} than BO. Likewise, in the 60D robotic task, PseudoBO - RP is 2.5X faster than BORE, and 4X than LFBO. Figure \ref{fig:pareto} shows a Pareto plot to illustrate the balance between the best objective value attained eventually and runtime. We see that PseudoBO variants yield better efficiency than BO, BORE and LFBO.

\begin{table}[htbp]
  \centering
  \scalebox{0.85}{
  \begin{tabular}{c|c|c|c|c}
    \toprule
     Tasks & Synthetic (10D) & NN Tuning (12D) & Robot Push (14D) & Rover (60D) \\
    
     Queries & 500 & 300 & 10,000 & 20,000 \\
    \midrule
    RS &  1.00  & 2.00 & 5.37 & 16.73 \\
    $\text{BO}^{*}$  & 399.90 & 791.40 & 6202.55 & - \\
    SMAC  & 747.4 & 200.14 & - & - \\
    TPE & 16.50  & 6.00  & 67.78 & - \\
    $\text{BORE}^{*}$ & 37.10 & 17.20  & 514.37 & 1361.19 \\
    $\text{LFBO}^{*}$ & 66.70 & 23.88 & 575.38 & 2130.78 \\
    PseudoBO - RP & 42.03 & 32.90 & 127.55 & 513.08 \\
    PseudoBO - LR + Hyb & 72.01 & 20.60  & 343.93 & 1453.00 \\
    \bottomrule
  \end{tabular}}
  \vspace{1mm}
  \caption{\small{The recorded runtime for all task types in seconds. A * on the top right of a method indicates that its implementation leverages GPU acceleration; otherwise, only CPUs are used. Even without GPU acceleration, PseudoBO variants demonstrate strong competitiveness in low-dimensional tasks (10D and 12D) with small query budgets and scale efficiently to high-dimensional tasks with larger query sizes, achieving shorter runtimes than other methods, except for TPE. Note that PseudoBO variants can be implemented in PyTorch to take advantage of GPU acceleration for further speed improvements.}}\label{tab: runtimes}
\end{table}

\section{Conclusion and Future Works}\label{sec:conclusion}

In this paper, we have re-evaluated the core principle of exploration-based black-box optimization. Our study is motivated by an apparent mismatch between BO theory and practice, in the sense that practical BO algorithms often lack theoretical guarantees offered by BO theory, which largely builds on GP-based procedures. To alleviate this mismatch, we propose a PseudoBO framework, which axiomatically dissects the minimal properties needed by the key ingredients in convergent black-box optimization algorithms. In particular, we show that if SP, UQ and AF satisfy their respective properties of local consistency, SNEB and improvement, then the assembled algorithm would eventually populate the decision space and consequently converges, and these properties can be further relaxed to achieve correspondingly relaxed consistency notions. Importantly, PseudoBO provides a recipe of algorithmic convergence beyond GP. That is, while GP is in the list of admissible ingredients, it is by no means the only option. From the recipe, we locate a combination of local regression as SP, a hybrid of randomized prior and minimum distance as UQ, and EI as AF, that gives rise to a consistent algorithm that performs competitively against existing state-of-the-art benchmarks across a wide range of tasks.

We view our work as a foundation for furthering the development of exploration-based black-box optimization that is beyond GP and theory-practice-balanced. Immediate extensions include the adaptation of our framework to handle noisy evaluations. In this situation, we face both epistemic uncertainty (due to lack of knowledge on the black-box objective function) and aleatory uncertainty (due to the evaluation noise). Much like the deterministic case, there has been works on extending GP-based algorithms to handle such problems, and our PseudoBO approach that dissects the key algorithmic ingredients could be generalized to tackle the additional aleatory uncertainty. Next is grey-box optimization. This consists of objective functions that can be composition of (multiple) black-box functions and analytically known functions. In this situation, we can individually model the SP and UQ for each underlying black-box functions and suitably aggregate them via a single AF, and achieve similar algorithmic consistency as in the full black-box case. A more challenging next step is to study rate results, but this likely requires an extent of opaque assumptions like in GP-based methods. Finally, an important direction is to provide a more principled approach to construct UQ, which is the core ingredient in inducing exploration, that can pair with specific SP in some arguably optimal sense, and also study the performance of such UQ relative to that of GP posterior.

\section*{Acknowledgements}
We gratefully acknowledge support from the Amazon CAIT Fellowship, as well as the InnoHK initiative, the Government of the HKSAR, and Laboratory for AI-Powered Financial Technologies. We also thank Peter Frazier and Wenjia Wang for the valuable comments and reference pointers that have greatly improved our work.

\bibliographystyle{abbrv}
\bibliography{references}

\newpage

{\LARGE{Appendices}}

\subsection{Proof of Theorem \ref{main basic}}
\begin{proof}{Proof.}
We first prove part 1 of the theorem. Denote $\{x_n\}$ as the sequence of evaluation points chosen by PseudoBO, and correspondingly $\mathcal D_n=\{(x_1,f(x_1)),\ldots,(x_n,f(x_n))\}$ is the data collection up to step $n$. Since $\mathcal X$ is compact, $\{x_n\}$ has a convergent subsequence, say $\{x_{\pi(n)}\}$. By Assumption \ref{basic assumptions} part 2, we have $W_n(x_{\pi(n)};\mathcal D_{\pi(n)-1})\to0$ and hence $\liminf_{{n}\to\infty}W_n(x_n;\mathcal D_{n-1})=0$. On the other hand, suppose there is a point, say $x'\in\mathcal X$, that is non-adherent to $\{x_n\}$, i.e., there is no subsequence in $\{x_n\}$ that converges to $x'$. Then, by Assumption \ref{basic assumptions} part 1, we have $\liminf_{{n}\to\infty}W_n(x';\mathcal D_n)> 0$. But this is impossible because PseudoBO requires at each step $x_{n+1}\in\text{argmax}_{x\in \X}W_n(x;\mathcal D_n)$, hence a contradiction.

Part 2 of the theorem follows by a straightforward use of the continuity of $f$. More precisely, as $\mathcal X$ is compact and $f$ is continuous, $x^*\in\text{argmax}_{x\in\mathcal X}f(x)$ is well-defined and there is a subsequence of $\{x_n\}$, say $\{x_{\pi'(n)}\}$, converging to $x^*$ by our first conclusion above. Then we have $f(x_{\pi'(n)})\to f(x^*)=Z^*$ by the continuity of $f$. This gives
\begin{equation*}
    Z^*\geq f(\hat x_n^*)\geq f(x_{\pi'(n)})\to f(x^*)=Z^*,
\end{equation*}
which concludes $f(\hat x_n^*)\to Z^*$.

\end{proof}

\subsection{Proof of Theorem \ref{PseudoBO main}}
\begin{proof}{Proof.}
Consider an arbitrary finite-cardinality set sequence $S_n\subset \X$ and $A_n=E_f(S_n)$. Suppose $\inf_n\Delta(x, S_{n}) > 0$. Then, by Assumption \ref{GNEB} part 1, we have $\liminf_{n\to\infty}\hat \sigma(x; A_{n})> 0$. By Assumption \ref{improvement} part 1, we must have $\liminf_{n\to\infty} W_n(x, A_{n})>0$. 

To verify Assumption \ref{basic assumptions} part 2, consider a sequence $x_n\in X$ that converges to $x'$, and $A_n=E_f(S_n)$ for an arbitrary finite-cardinality set sequence $S_n\subset X$. By Assumption \ref{GNEB} part 2, we have $\hat\sigma(x_n;A_{n-1}\cup \D_{n-1})\to0$ where $\D_{n-1}=\{(x_1,f(x_1)),\ldots,(x_{n-1},f(x_{n-1}))\}$. Moreover, by Assumption \ref{local consistency} we have $\hat f(x_n;A_{n-1}\cup \D_{n-1})\to f(x')$. Thus, 
\begin{equation*}
    \zeta(\hat f(x_n;A_{n-1}\cup \D_{n-1})-\max\Pi_f(A_{n-1}\cup \D_{n-1}))\leq \zeta(\hat f(x_n;A_{n-1}\cup \D_{n-1}) - f(x_{n-1})) \to \zeta(0) \leq 0,
\end{equation*}
by the assumed continuity of $f$ and $\zeta$, the non-decreasing property of $\zeta$ and $\zeta(0) \leq 0$. Thus, by Assumption \ref{improvement} part 2, we further have 
\begin{equation*}
    W_n(x_n; A_{n-1}\cup \D_{n-1}) = g_n(\zeta(\hat f(x_n; A_{n-1}\cup \D_{n-1}) - \max\Pi_f(A_{n-1}\cup \D_{n-1})), \hat\sigma(x_n;A_{n-1}\cup \D_{n-1}))\to0.
\end{equation*}
This concludes Assumption \ref{basic assumptions} part 2.

\end{proof}

\subsection{Proof of Theorem \ref{main basic relaxed}}
\begin{proof}{Proof.}
We first prove part 1 of the theorem. Denote $\{x_n\}$ as the sequence of evaluation points chosen by PseudoBO, and correspondingly $\mathcal D_n=\{(x_1,f(x_1)),\ldots,(x_n,f(x_n))\}$ is the data collection up to step $n$. Since $\mathcal X$ is compact, $\{x_n\}$ has a convergent subsequence, say $\{x_{\pi(n)}\}$. By Assumption \ref{basic assumptions relaxed} part 2, we have $W_n(x_{\pi(n)};\mathcal D_{\pi(n)-1})\to0$ and hence $\liminf_{{n}\to\infty}W_n(x_n;\mathcal D_{n-1})=0$. On the other hand, suppose there is a point, say $x'\in\mathcal X$, that satisfies $\inf_n\Delta(x',X_n)>\delta$. Then, by Assumption \ref{basic assumptions relaxed} part 1, we have $\liminf_{{n}\to\infty}W_n(x';\mathcal D_n)> 0$. But this is impossible because PseudoBO requires at each step $x_{n+1}\in\text{argmax}_{x\in \X}W_n(x;\mathcal D_n)$, hence a contradiction.

Now we prove part 2 of the theorem. First, as $\mathcal X$ is compact and $f$ is continuous, $x^*\in\text{argmax}_{x\in\mathcal X}f(x)$ is well-defined. By our first conclusion above, for any $\eta>0$, we can find an $x_n$ such that
$\|x_n-x^*\|\leq\delta+\eta$. Then we have 
\begin{equation*}
    f(\hat x_n^*)\geq f(x_n)\geq f(x^*)-L(\delta+\eta),
\end{equation*}
by the Lipschitzness of $f$. Since $\eta$ is arbitrary, we have
$$\liminf_nf(\hat x_n^*)\geq f(x^*)-L\delta=Z^*-L\delta$$
which concludes the theorem.

\end{proof}

\subsection{Proof of Theorem \ref{PseudoBO main relaxed}}
\begin{proof}{Proof.}
Consider an arbitrary finite-cardinality set sequence $S_n\subset \X$, and $A_n=E_f(S_n)$. Suppose $\inf_n\Delta(x, S_{n}) > \delta$. Then, by Assumption \ref{GNEB relaxed} part 1, we have $\liminf_{n\to\infty}\hat \sigma(x; A_{n})> 0$. By Assumption \ref{improvement} part 1, we must have $\liminf_{n\to\infty} W_n(x, A_{n})>0$. 

To verify Assumption \ref{basic assumptions relaxed} part 2, consider a sequence $x_n\in X$ that converges to $x'$, and $A_n=E_f(S_n)$ for an arbitrary finite-cardinality set sequence $S_n\subset X$. By Assumption \ref{GNEB relaxed} part 2, we have $\hat\sigma(x_n;A_{n-1}\cup \D_{n-1})\to0$ where $\D_{n-1}=\{(x_1,f(x_1)),\ldots,(x_{n-1},f(x_{n-1}))\}$. Moreover, by Assumption \ref{local consistency relaxed} we have $\limsup_n\hat f(x_n;A_{n-1}\cup \D_{n-1})\leq f(x')+\epsilon$. Thus, 
\begin{equation*}
    \zeta(\hat f(x_n;A_{n-1}\cup \D_{n-1})-\max\Pi_f(A_{n-1}\cup \D_{n-1})-\epsilon)\leq \zeta(\hat f(x_n;A_{n-1}\cup \D_{n-1}) - f(x_{n-1})-\epsilon)
\end{equation*}
and so
\begin{align*}
&\hspace{5mm}\limsup_n\zeta(\hat f(x_n;A_{n-1}\cup \D_{n-1})-\max\Pi_f(A_{n-1}\cup \D_{n-1})-\epsilon)\\
&\leq \zeta(\limsup_n\{\hat f(x_n;A_{n-1}\cup \D_{n-1}) - f(x_{n-1})-\epsilon\})\\
    &\leq \zeta(f(x')+\epsilon - f(x^*)-\epsilon)\\
    &\leq \zeta(0)\\
    &\leq 0
\end{align*}
by the assumed continuity of $f$ and $\zeta$, the non-decreasing property of $\zeta$ and $\zeta(0) \leq 0$. Thus, by Assumption \ref{improvement} part 2, we further have 
\begin{equation*}
    W_n(x_n; A_{n-1}\cup \D_{n-1}) = g_n(\zeta(\hat f(x_n; A_{n-1}\cup \D_{n-1}) - \max\Pi_f(A_{n-1}\cup \D_{n-1})-\epsilon), \hat\sigma(x_n;A_{n-1}\cup \D_{n-1}))\to0.
\end{equation*}
This concludes Assumption \ref{basic assumptions relaxed} part 2.

\end{proof}

\subsection{Proof of Proposition \ref{prop:GP local consistent}}
\begin{proof}{Proof.}
Proposition 10 in \cite{vazquez2010convergence} assumes that $f \in \mathcal{H}$, where $\mathcal{H}$ denotes the RKHS of $\K$. Consider two sequences $\{x_n\}$, $\{y_n\} \subset \X$ that is compact. Suppose $y_n \rightarrow y'$ such that $y'$ is adherent to $\{x_n\}$. This is the condition (i) in Proposition 10 of \cite{vazquez2010convergence}.  A partial result of Proposition 10 demonstrates that under this condition, GP mean predictor $\hat{f}_{GP}(y_n; \D_n) \rightarrow f(y')$.

To further show it satisfying our local consistency assumption, consider a sequence $x_n\in \X$ that converges to $x'$, and $A_n=E_f(S_n)$ for an arbitrary finite-cardinality set sequence $S_n\subset \X$. By definition, $x'$ is an adherent point of $S_n\cup X_n$. Therefore, we have $\hat f(x_n;A_{n-1}\cup \D_{n-1})\to f(x')$.

\end{proof}

\subsection{Proof of Proposition \ref{prop:Nearest local consistent}}
\begin{proof}{Proof.}
Suppose $\{x_{n}\}$ is a sequence in $\X$ that converges to $x'$, and $A_n=E_f(S_n)$ for an arbitrary finite-cardinality set sequence $S_n\subset \X$. Define $x'_n := \argmin_{y\in X_{n-1}\cup S_{n-1}} ||x_n - y||$ where $X_{n-1}=\{x_1,\ldots,x_{n-1}\}$. We have $\hat{f}_{\text{NN}}(x_n;A_{n-1}\cup\D_{n-1}) = f(x'_n)$. Note that $x'_n\to x'$. Thus, $\hat{f}_{\text{NN}}(x_n;A_{n-1}\cup\D_{n-1}) = f(x'_n) \to f(x')$ by the continuity of $f$.

\end{proof}
 

\subsection{Proof of Proposition \ref{prop:ONN local consistent}}
\begin{proof}{Proof.}
Suppose $\{x_{n}\}$ is a sequence in $\X$ that converges to $x'$, and $A_n=E_f(S_n)$ for an arbitrary finite-cardinality set sequence $S_n\subset \X$. We have
\begin{align}
&|\hat f_{\text{Net}}(x_n; A_{n-1}\cup \D_{n-1}) - \hat f_{\text{Net}}(x_{n-1}; A_{n-1}\cup \D_{n-1})|\notag\\
&\leq L\norm{x_n - x_{n-1}}\text{\ \ \ \ since $\hat{f}_{\text{Net}}$ is $L-$Lipschitz}\notag\\
&\to 0\text{\ \ \ \ by the convergence and hence Cauchy property of $\{x_n\}$}\label{eq:ONN convergence}
\end{align}
Thus, 
\begin{align*}
&|\hat f_{\text{Net}}(x_n; A_{n-1}\cup \D_{n-1}) - f(x')|\\
&\leq|\hat f_{\text{Net}}(x_n; A_{n-1}\cup \D_{n-1}) - \hat f_{\text{Net}}(x_{n-1}; A_{n-1}\cup \D_{n-1})|+|\hat f_{\text{Net}}(x_{n-1}; A_{n-1}\cup \D_{n-1})-f(x')|\\
&\to 0
\end{align*}
by the triangle inequality, \eqref{eq:ONN convergence} and $\hat f_{\text{Net}}(x_{n-1}; A_{n-1}\cup \D_{n-1})=f(x')$ thanks to the over-parametrization.

\end{proof}

\subsection{Proof of Proposition \ref{prop:Tree local consistent}}
\begin{proof}{Proof.}
Suppose $x_n$ is a sequence in $\X$ that converges to $x'$, and $A_n = E_f(S_n)$ for an arbitrary finite-cardinality set sequence $S_n \in \X$. Then there exists $N>0$ such that for all $n\geq N$, we have $\|x_n - x'\|\leq \text{diam}(R_{l(x', \theta)})$. For $n\geq N$, we have
\begin{align*}
    |\hat{f}_{\text{tree}}(x_n ; A_{n-1}\cup \D_{n-1}) - f(x')| &= |\sum_{i=1}^{n-1} \frac{\mathbbm{1}\{x_i \in R_{l(x_n, \theta)}\}}{\sum_{i=1}^{n-1}\mathbbm{1}\{x_i \in R_{l(x_n, \theta)}\}}f(x_i) - f(x')|\\
    &=|\frac{1}{|\mathcal{I}_{n-1}(R_{l(x_n, \theta)})|} \sum_{x_i \in \mathcal{I}_{n-1}(R_{l(x_n, \theta)})} f(x_i) - f(x')|\\
    &\leq \frac{1}{|\mathcal{I}_{n-1}(R_{l(x_n, \theta)})|} \sum_{x_i \in \mathcal{I}_{n-1}(R_{l(x_n, \theta)})} |f(x_i) - f(x')|\\
    &\leq \frac{1}{|\mathcal{I}_{n-1}(R_{l(x_n, \theta)})|} \sum_{x_i \in \mathcal{I}_{n-1}(R_{l(x_n, \theta)})} L\|x_i-x'\|\\
    &\leq \frac{1}{|\mathcal{I}_{n-1}(R_{l(x_n, \theta)})|} \sum_{i \in \mathcal{I}_{n-1}(R_{l(x_n, \theta)})} L\cdot(\text{diam}(R_{l(x_n; \theta)}+\text{diam}(R_{l(x'; \theta)}))\\
    &= 2L\cdot\max_{x\in\X}\text{diam}(R_{l(x; \theta)}),
\end{align*}
where $|\mathcal{I}_{n-1}(R_{l(x_n, \theta)})|$ represents the cardinality of the set $\mathcal{I}_{n-1}(R_{l(x_n, \theta)})$ comprising all points in $A_{n-1}\cup\D_{n-1}$ located inside the region $R_{l(x_n, \theta)}$, the last but one inequality follows by the $L$-Lipschitz property, and the last inequality follows from the triangle inequality.

\end{proof}

\subsection{Proof of Proposition \ref{prop:KR local consistent}}
\begin{proof}{Proof.}
Suppose $x_n$ is a sequence in $\X$ that converges to $x'$, and $A_n=E_f(S_n)$ for an arbitrary finite-cardinality set sequence $S_n\subset \X$. Hence $x_n$ is Cauchy. So, for any $n$ sufficiently large, we have $\|x_n-x_{n-1}\|\leq Mh$. For any such $n$, we have
\begin{align*}
\hat f_{\text{LR}}(x_n;A_{n-1}\cup\D_{n-1})&=\frac{\sum_{i\leq n-1}K\left(\frac{\|x_n-x_i\|}{h}\right)f(x_i)}{\sum_{i\leq n-1}K\left(\frac{\|x_n-x_i\|}{h}\right)}\\
&=\frac{\sum_{i\leq n-1}K\left(\frac{\|x_n-x_i\|}{h}\right)(f(x_n)+e_n)}{\sum_{i\leq n-1}K\left(\frac{\|x_n-x_i\|}{h}\right)}
\end{align*}
where $|e_n|\leq L\|x_i-x_n\|$ by the $L$-Lipschitzness of $f$. Thus, the above is equal to $f(x_n)+\tilde e_n$ where $|\tilde e_n|\leq LMh$ since $K$ has support $[0,M]$.

So, we have
\begin{align*}
& \limsup_n|\hat f_{\text{LR}}(x_n;A_{n-1}\cup\D_{n-1})-f(x')|\\
&\leq \limsup_n|\hat f_{\text{LR}}(x_n;A_{n-1}\cup\D_{n-1})-f(x_n)|+\lim_n|f(x_n)-f(x')|\\
&\leq LMh
\end{align*}
which concludes the proposition.

\end{proof}

\subsection{Proof of Proposition \ref{prop:RP local consistent}}
\begin{proof}{Proof.}
Suppose $\{x_{n}\}$ converges to $x'$, and $A_n=E_f(S_n)$ for an arbitrary finite-cardinality set sequence $S_n\subset \X$. For each $r \sim \mathcal{R}$, by $\epsilon$-relaxed local consistency and the continuity of $r$, we have
\begin{align}
    \hat f^{(r)}(x_n;A_{n-1}\cup\D_{n-1}) &= r(x_n) + \hat{f}(x_n; A^{(r)}_{n-1}\cup\D^{(r)}_{n-1})\nonumber\\
    &= r(x') + f(x') - r(x') +e_n=f(x')+e_n\label{eq:rp sample consist}
\end{align}
where $\limsup_n|e_n|\leq \epsilon$. 
Thus, since $f,\hat f^{(r)}$ are bounded, we have
\begin{equation*}
  \limsup_n|E_{r\sim\mathcal R}[\hat{f}^{(r)}(x_n ;A_{n-1}\cup\D_{n-1})]-f(x')|\leq E_{r\sim\mathcal R}[\limsup_n|\hat{f}^{(r)}(x_n ;A_{n-1}\cup\D_{n-1})-f(x')|] \leq\epsilon.  
\end{equation*}

\end{proof}


\subsection{Proof of Proposition \ref{prop:Hybrid local consistent}}
\begin{proof}{Proof.}
Suppose $\{x_{n}\}$ converges to $x'$, and $A_n=E_f(S_n)$ for an arbitrary finite-cardinality set sequence $S_n\subset \X$. Given a finite class of $\epsilon$-relaxed locally consistent SPs $\hat{\mathcal F}=\{\hat f_i\}$, we have
\begin{equation*}
    \limsup_n\left|\sum_{i = 1}^{|\hat{\mathcal{F}}|} \alpha_i\hat f_i(x_n; A_{n-1} \cup \D_{n-1})-f(x')\right|\leq  \sum_{i = 1}^{|\hat{\mathcal{F}}|} \alpha_i \limsup_n| \hat f_i(x_n; A_{n-1} \cup \D_{n-1})-f(x')|\leq \epsilon,
\end{equation*}
with the assumption that $\sum_{i = 1}^{|\hat{\mathcal{F}}|}\alpha_i = 1$.

\end{proof}



Recall the NEB property of GP in \cite{vazquez2010convergence}:
\begin{assumption}[NEB property]
A GP has the NEB property if, for all sequence $\{x_n\}$ in $\X$ and all $x'$ in $\X$, the following statements are equivalent:
\begin{enumerate}
\item $x'$ is an adherent point of $\{x_n\}$, i.e., there is a subsequence in $\{x_n\}$ that converges to $x'$.
\item $\hat{\sigma}_{\text{GP}}^2 (x'; \D_n) \to 0$ as $n\to\infty$.
\end{enumerate}\label{NEB}
\end{assumption}
Here $\hat{\sigma}_{\text{GP}}^2 (x'; \D_n)$ denotes the posterior variance of the GP with data $\D_n$ collected. We are ready to prove Proposition \ref{prop:GP GNEB}:

\subsection{Proof of Proposition \ref{prop:GP GNEB}}
\begin{proof}{Proof.} 
Proposition 10 in \cite{vazquez2010convergence} verifies the NEB property of GP, under the assumptions that $\X$ is compact, the GP is centered, the kernel function $\mathcal{K}$ of the GP is continuous and stationary, and its spectral density $\mathcal{S}$ satisfying that $\mathcal{S}^{-1}$ is at most polynomial growth. We show our SNEB property below based on this result. Moreover, clearly it is equivalent to show the required properties for the posterior variance $\hat{\sigma}_{\text{GP}}^2 (x'; \D_n)$ as the posterior standard deviation $\hat{\sigma}_{\text{GP}} (x'; \D_n)$.

To show part 1 of Assumption \ref{GNEB}, suppose $\inf_n \Delta(x; S_n) > 0$ for an arbitrary finite-cardinality set sequence $S_n\subset\X$. This implies $\inf_n \Delta(x; \cup_{k\leq n}S_k) > 0$. Therefore, $x$ cannot be an adherent point of $\cup_{k\leq n}S_k$. The implication of Assumption \ref{NEB} from statement 2 to statement 1 tells us $\hat{\sigma}_{\text{GP}}^2 (x'; \cup_{k\leq n}S_{k}) \not\to 0$, and since $\hat{\sigma}_{\text{GP}}^2 (x'; \cup_{k\leq n}S_{k}) \geq 0$ and is monotonically non-increasing on $n$ by the non-decreasing property of $\cup_{k\leq n}S_{k}$, we must have $\liminf_{n\to\infty}\hat{\sigma}_{\text{GP}}^2 (x'; S_{n})\geq\liminf_{n\to\infty}\hat{\sigma}_{\text{GP}}^2 (x'; \cup_{k\leq n}S_{k}) > 0$.

Denote the GP as $\xi(\cdot)$. To show part 2 of Assumption \ref{GNEB}, suppose $x_{n}$ converges to $x'$, and $A_n=E_f(S_n)$ for an arbitrary finite-cardinality set sequence $S_n\subset \X$. We have
\begin{align*}
    \hat{\sigma}_{\text{GP}}^2 (x_n; A_{n-1}\cup\D_{n-1}) &\labelrel\leq{sign:GP NEB 1}  \hat{\sigma}_{\text{GP}}^2 (x_n; \D_{n-1}) \\
    &\labelrel={sign:GP NEB 2} \Var[\xi(x_n) - \hat{f}_{\text{GP}}(x_n; \D_{n-1})] \\
    &\labelrel\leq{sign:GP NEB 3} \Var[\xi(x_n) - \xi(x_{n-1})]\\
    &\labelrel={sign:GP NEB 4}  \mathcal{K}(x_n, x_n) - 2\mathcal{K}(x_n, x_{n-1}) + \mathcal{K}(x_{n-1}, x_{n-1}) \labelrel\to{sign:GP NEB 5}  0,
\end{align*}
where $\hat{f}_{\text{GP}}$ is the posterior mean of the GP, and inequality~\eqref{sign:GP NEB 1} follows from the non-increasing property of the posterior variance as more points are sampled; the equality~\eqref{sign:GP NEB 2} follows by the definition of variance; the inequality~\eqref{sign:GP NEB 3} follows by the fact that the posterior mean is the $\mathcal L^2$ best linear predictor; the inequality~\eqref{sign:GP NEB 4} follows by expanding out the variance; and the convergence of~\eqref{sign:GP NEB 5} follows by the continuity of $\mathcal{K}$ and the convergence and hence Cauchyness of $x_n$.

\end{proof}


\subsection{Proof of Proposition \ref{prop:MD GNEB}}
\begin{proof}{Proof.}
To verify part 1 of Assumption \ref{GNEB}, suppose $\inf_n \Delta(x; S_n \cup X_n) > 0$. Then $\liminf_{n\to\infty} \Delta(x; S_n \cup X_n) \geq \inf_n \Delta(x; S_n \cup X_n) > 0$.

To verify part 2 of Assumption \ref{GNEB}, suppose $x_n\to x'$ for some $x'$. We have
$\Delta(x_n; S_{n-1}\cup X_{n-1}) \leq ||x_n - x_{n-1}||\to0$
since $x_n$ is Cauchy. 

\end{proof}

\subsection{Proof of Proposition \ref{prop:RP GNEB}}
\begin{proof}{Proof.}
Part 1 of Assumption \ref{GNEB relaxed} follows directly from the assumption of the proposition. To verify part 2, suppose $\{x_{n}\}$ converges to $x'$, and $A_n=E_f(S_n)$ for an arbitrary finite-cardinality set sequence $S_n\subset \X$. From $\epsilon$-relaxed local consistency of $\hat f^{(r)}$ and the continuity of $r$, we have $\limsup_n|\hat f^{(r)}(x_n;A_{n-1}\cup\D_{n-1})-f(x')|\leq\epsilon$
for any $r$ almost surely by following the first part of the proof of Proposition \ref{prop:RP local consistent}. Then, since $\hat f^{(r)}$ is uniformly bounded, we have
\begin{align*}
\limsup_n(Var_{r\sim\mathcal R}[\hat f^{(r)}(x_n;A_{n-1}\cup\D_{n-1})])^{\nicefrac{1}{2}}&\leq\limsup_n(E_{r\sim\mathcal R}[\hat f^{(r)}(x_n;A_{n-1}\cup\D_{n-1})^2])^{\nicefrac{1}{2}}\\
&\leq(E_{r\sim\mathcal R}[\limsup_n\hat f^{(r)}(x_n;A_{n-1}\cup\D_{n-1})^2])^{\nicefrac{1}{2}}\\
&\leq\epsilon
\end{align*}
and so
$$((Var_{r\sim\mathcal R}[\hat f^{(r)}(x_n;A_{n-1}\cup\D_{n-1})])^{\nicefrac{1}{2}}-\epsilon)_+\to0$$

\end{proof}

\subsection{Proof of Corollary \ref{prop:RP GNEB kernel}}
\begin{proof}{Proof.}
To apply Proposition \ref{prop:RP GNEB}, we verify the assumptions in Proposition \ref{prop:RP local consistent} for the local regression SP, with $\epsilon=(L+\tilde L)Mh$. To this end, note that with the assumptions in Proposition \ref{prop:KR local consistent} and the additional assumptions in Corollary \ref{prop:RP GNEB kernel}, $f-r$ is an $(L+\tilde L)$-Lipschitz function. Thus, together with the continuity of $r$, $\hat f^{(r)}$ is $(L+\tilde L)Mh$-relaxed locally consistent for almost surely any $r$. Moreover, the uniform boundedness of $r$ and $f$ also implies the same property for $\hat f^{(r)}$ driven by the local regression SP. These verify all the assumptions in Proposition \ref{prop:RP local consistent}.

Next, note that when $\inf_n\Delta(x_0;\D_n)>Mh$, we have $\hat f_{\text{Ker}}(x_0;\D_n^{(r)})$ equal to a prefixed constant by the construction of the local regression SP. In this case, $\hat f^{(r)}(x_0;\D_n)=r(x_0)$. So, by assumption we have
$(\Var_{r\sim \mathcal{R}} [\hat{f}^{(r)}(x_0; \D_n)])^{\nicefrac{1}{2}}=(\Var_{r\sim\mathcal R}[r(x_0)])^{\nicefrac{1}{2}}>(L+\tilde L)Mh$. This verifies the last assumption in Proposition \ref{prop:RP GNEB} with $\delta=Mh$. Hence we conclude the corollary.

\end{proof}

\subsection{Proof of Proposition \ref{prop: hybrid UQ GNEB}}
\begin{proof}{Proof.}
Given a class of UQs $\hat{\Sigma} := \{ \hat{\sigma_1}, \hat{\sigma_2}, ...\}$ with the $\delta$-relaxed SNEB property, to show part 1 of Assumption \ref{GNEB relaxed}, suppose $\inf_n \Delta(x; S_n) > \delta$, with $A_n=E_f(S_n)$ for an arbitrary finite-cardinality set sequence $S_n\subset \X$. For any convex combination of $\hat{\Sigma}$,
\begin{equation*}
\liminf_{n\to\infty} \sum_{i = 1}^{|\hat{\Sigma}|} \alpha_i\hat{\sigma}_i(x; A_{n-1} \cup \D_{n-1})=  \sum_{i = 1}^{|\hat{\Sigma}|} \alpha_i \liminf_{n\to\infty} \hat{\sigma}_i(x; A_{n-1} \cup \D_{n-1}) > 0,
\end{equation*}   
where the last inequality is inherited from part 1 of Assumption \ref{GNEB relaxed} for the individual UQs and that at least one of $\alpha_i$'s is positive since they sum up to 1.

To show part 2 of Assumption \ref{GNEB relaxed}, suppose $\{x_{n}\}$ converges to $x'$, and $A_n=E_f(S_n)$ for an arbitrary finite-cardinality set sequence $S_n\subset \X$. For any convex combination of $\hat{\Sigma}$,
\begin{equation*}
    \lim_{n\to\infty} \sum_{i = 1}^{|\hat{\Sigma}|} \alpha_i\hat{\sigma}_i(x_n; A_{n-1} \cup \D_{n-1})
    =  \sum_{i = 1}^{|\hat{\Sigma}|} \alpha_i \lim_{n\to\infty} \hat{\sigma}_i(x_n; A_{n-1} \cup \D_{n-1}) = 0.
\end{equation*} 

\end{proof}





\subsection{Proof of Proposition \ref{prop: PI improve}}
\begin{proof}{Proof.}

To verify part 1 of Assumption \ref{improvement}, suppose $\liminf_{n\to\infty} p_n > -\infty$ and $\liminf_{n\to\infty}q_n > 0$. Then, for $n\geq N$ for some large $N$, we have $p_n\geq c_1>-\infty$ and $q_n\geq c_2>0$ for some $c_1$ and $c_2$, and so $g_n^{\text{PI}}(p_n,q_n)\geq\inf_{n\geq N}\Phi((c_1-\tau)/q_n)$ which is at least $\Phi((c_1-\tau)/c_2)$ if $c_1-\tau<0$, and $1/2$ if $c_1-\tau\geq0$. Thus, $\liminf_{n\to\infty} g_n^{\text{PI}}(p_n,q_n)>0$.

To verify part 2 of Assumption \ref{improvement}, suppose $\limsup_{n\to\infty}p_n \leq0$ and $q_n\to0$. Then $\limsup_{n\to\infty}p_n-\tau <0$ since $\tau>0$ and thus eventually $p_n-\tau\leq c<0$ for some $c$, giving $g_n^{\text{PI}}(p_n,q_n)\to0$ by directly using \eqref{EI def}.
\end{proof}

\begin{proof}[Proof of Proposition \ref{prop: EI improve}]
To verify part 1 of Assumption \ref{improvement}, suppose $\liminf_{n\to\infty} p_n > -\infty$ and $\liminf_{n\to\infty}q_n>0$. Then, for $n\geq N$ for a large enough $N$, we have $p_n\geq c_1>-\infty$ and $q_n\geq c_2>0$ for some $c_1,c_2$. Note that $g_n^{\text{EI}}(p_n,q_n)=E(N(p_n,q_n)-\tau)_+$, where $N(p_n,q_n)$ denotes a normal variable with mean $p_n$ and standard deviation $q_n$. Suppose $n\geq N$. We have $N(p_n,q_n)$ stochastically dominates $N(c_1,q_n)$, and thus $E(N(p_n,q_n)-\tau)_+\geq E(N(c_1,q_n)-\tau)_+$ since $(\cdot-\tau)_+$ is a non-decreasing. Moreover, we have $N(c_1,q_n)$ second-order stochastically dominates $N(c_1,c_2)$, and thus $E(N(c_1,q_n)-\tau)_+\geq E(N(c_1,c_2)-\tau)_+$ since $(\cdot-\tau)_+$ is non-decreasing and convex. Hence $g_n^{\text{EI}}(p_n,q_n)\geq g_n^{\text{EI}}(c_1,c_2)>0$. This gives $\liminf_{n\to\infty} g_n(p_n,q_n)>0$.

To verify part 2 of Assumption \ref{improvement}, suppose $\limsup_{n\to\infty}p_n \leq0$ and $q_n\to0$. Then $g_n^{\text{EI}}(p_n,q_n)\to0$ by directly using \eqref{EI def} and noting that the function is continuous even at $q_n=0$.

\end{proof}

\subsection{Proof of Proposition \ref{prop: UCB+ improve}}
\begin{proof}{Proof.}
Since $\tau$ is fixed, $\beta_n\to\infty$ and $p_n$ is bounded, $\frac{p_n - \tau}{\beta_n}\to0$. To verify part 1 of Assumption \ref{improvement}, supposing $\liminf_{n\to\infty}q_n>0$, we have $\liminf_{n\to\infty}g_n^{\text{UCB}}(p_n, q_n) = \liminf_{n\to\infty}\{\frac{p_n - \tau}{\beta_n} + q_n\}=\liminf_{n\to\infty}q_n>0$. To verify part 2, supposing $q_n\to0$, we have $g_n^{\text{UCB}}(p_n, q_n) = \frac{p_n - \tau}{\beta_n} + q_n\to0$. 

\end{proof}

\subsection{Proof of Proposition \ref{prop: hybrid improve}}
\begin{proof}{Proof.}
Both parts of Assumption \ref{improvement} can be straightforwardly shown to preserve under convex combinations.

\end{proof}

\section{Additional Experiment Details}\label{appendix: exp details}

\subsection{Implementation Details}

The methods we compare with are listed in Table \ref{tab: implementations}. 

\begin{table}[htbp]
  \centering
  \scalebox{1}{
  \begin{tabular}{ccc}
    \toprule

    Methods & Software library & URL \\
    \midrule
    RS & Hyperopt & https://github.com/hyperopt/hyperopt \\
    
    BO & BoTorch & https://botorch.org \\
    
    
    TPE & Hyperopt  & https://github.com/hyperopt/hyperopt \\

    BORE & Syne Tune  & https://github.com/awslabs/syne-tune \\

    LFBO & -  & https://github.com/lfbo-ml/lfbo \\
    \bottomrule
  \end{tabular}}
  \vspace{1mm}
  \caption{\small{Package information.}}\label{tab: implementations}
\end{table}

In particular, for standard BO, we use Matérn Kernel with the default hyperparameter values in BoTorch. For BORE and LFBO, we use XGBoost as the classifier with preset parameters in its original implementations, since LFBO and BORE with XGBoost has relatively good and stable performances across tasks and are computationally much faster than other classifiers (e.g., random forest or neural network). 

All experiments are conducted on a computer with a 4-core Intel(R) Xeon(R) CPU @ 2.30GHz and a Tesla T4 GPU.

\subsubsection{Hyperparameters for PseudoBO Methods}

\paragraph{Randomized prior functions.}The random functions in all tasks are sampled from a random 3-layer neural network $r(x) = W_3 \text{Tanh}(W_2 \text{Tanh}(W_1 x + b_1) + b_2) + b_3$, with the Glorot random initialization \cite{glorot2010understanding}.

\paragraph{Kernel.}We use Gaussian kernel throughout all models and all tasks.

\paragraph{Hybrid weight.}We fix $\alpha$ to be 0.95 in all tasks. 

\paragraph{Bandwidths.}PseudoBO - RP uses $0.1\times(\X_{i,1} - \X_{i,0})$ for synthetic functions optimization; $h_{0}' = 0.1 \times [1/2, 1/2, 1/2, 1/2, 1/4, 1/3, 1/3, 1/6, 1/2, 1/2, 1/6, 1/6]$ for neural network tuning, where the denominators in the vector represents the number of candidates in each hyperparameter (see Section \ref{subsec: tuning detail} for further details);  $h_{0,i}' = 0.1(\X_{i,1} - \X_{i,0})$ for both robotic tasks.

PseudoBO - LR + Hyb uses $0.01 \times [1/2, 1/2, 1/2, 1/2, 1/4, /3, 1/3, 1/6, 1/2, 1/2, 1/6, 1/6]$ for neural network tuning, and $0.001 \times (\X_{i,1} - \X_{i,0})$ for all other tasks for UQ. For SP, we use $0.1\times (\X_{i,1} - \X_{i,0})$, $h_{0,i}^{(u)} = 0.2(\X_{i,1} - \X_{i,0})$ for synthetic functions optimization; $0.75\times [1/2, 1/2, 1/2, 1/2, 1/4, /3, 1/3, 1/6, 1/2, 1/2, 1/6, 1/6]$ for neural network tuning; $0.1\times(\X_{i,1} - \X_{i,0})$ for robot pushing; $0.2\times (\X_{i,1} - \X_{i,0})$ for rover trajectory planning.

\paragraph{Perturbation probability of Sobol sequence.}As in \cite{eriksson2019scalable}, we use the Sobol sequence for inner optimization of all PseudoBO methods, with perturbing probability of $1$ in the 2D synthetic function tasks, $0.75$ in the 6D synthetic function task, $0.5$ in the 10D synthetic function task, $0.4$ in the 12D neural network tuning task, $0.35$ in the 14D robot push task, and $0.15$ in the 60D rover trajectory task.


\subsection{Calibrated Coverage Rate}\label{appendix: CCR_details}
We employ our proposed CCR criterion to assess how well the UQs of the considered methods, GP, NN + MD, RP and LR + Hyb, are calibrated. We generate the training set $\D_{\text{train}}$, the validation set $\D_{\text{val}}$ and the test set $\D_{\text{train}}$ by uniformly sampling $20$, $10$ and $150$ points from the decision space at random, with their labels evaluated by the black-box function. 

We have purposefully designed the sizes of $\D_{\text{train}}$, $\D_{\text{val}}$, and $\D_{\text{test}}$. The size of the validation set is smaller than the size of training set so that the calibration over the validation set does not disclose excessive information about the unknown function's shape. Consequently, the quality of calibration depends on a combination of learning from the training set and the supplementary information from the validation set. Moreover, the size of test set is much larger than both for the purpose of a more accurate evaluation of the true performance of each UQ method.

To find $\lambda_{\text{val}} = \min_{\lambda \geq 0} \lambda$ such that 
\begin{equation*}
   \pr_{(x, y)\sim\D_{\text{val}}}(x \in [\hat{f}(x; \D_{\text{train}}) - \lambda\hat{\sigma}(x; \D_{\text{train}}), \hat{f}(x; \D_{\text{train}}) + \lambda\hat{\sigma}(x; \D_{\text{train}})]) = 1,
\end{equation*}
we use the bisection approach in Algorithm \ref{algorithm_calibrate}:

\begin{algorithm}
\caption{Pre-trained SP and UQ Combo Calibration}
\label{algorithm_calibrate}

\begin{algorithmic}[1]  
\Require Pretrained SP $\hat{f}(\cdot; \mathcal{D}_{\text{train}})$, pretrained UQ $\hat{\sigma}(\cdot; \mathcal{D}_{\text{train}})$, 
validation set $\mathcal{D}_{\text{val}}$, and tolerance level $\epsilon$
\Ensure Multiplier $\lambda_{\text{val}}$

\State Initialize $\lambda_{l} \gets 0$, $\lambda_{\text{init}} \gets 1$, $\lambda_{u} \gets \infty$

\While{$\lambda_{u} = \infty$}
    \State Compute $CR_{\text{init}} \gets \Pr_{(x, y) \sim \mathcal{D}_{\text{val}}}\left(x \in [\hat{f}(x; \mathcal{D}_{\text{train}}) 
    - \lambda_{\text{init}} \hat{\sigma}(x; \mathcal{D}_{\text{train}}), \hat{f}(x; \mathcal{D}_{\text{train}}) 
    + \lambda_{\text{init}} \hat{\sigma}(x; \mathcal{D}_{\text{train}})]\right)$

    \If{$CR_{\text{init}} < 1$}
        \State $\lambda_{\text{init}} \gets 2 \cdot \lambda_{\text{init}}$
    \Else
        \State $\lambda_{u} \gets \lambda_{\text{init}}$
    \EndIf
\EndWhile

\While{$\lambda_{u} - \lambda_{l} > \epsilon$}
    \State $\lambda_{\text{val}} \gets (\lambda_{l} + \lambda_{u}) / 2$
    
    \State Compute $CR_{\text{val}} \gets \Pr_{(x, y) \sim \mathcal{D}_{\text{val}}}\left(x \in [\hat{f}(x; \mathcal{D}_{\text{train}}) 
    - \lambda_{\text{val}} \hat{\sigma}(x; \mathcal{D}_{\text{train}}), \hat{f}(x; \mathcal{D}_{\text{train}}) 
    + \lambda_{\text{val}} \hat{\sigma}(x; \mathcal{D}_{\text{train}})]\right)$

    \If{$CR_{\text{val}} < 1$}
        \State $\lambda_{l} \gets \lambda_{\text{val}}$
    \Else
        \State $\lambda_{u} \gets \lambda_{\text{val}}$
    \EndIf
\EndWhile

\end{algorithmic}
\end{algorithm}

\begin{figure}[htbp!]
    \centering
    \begin{subfigure}[t]{\textwidth}
        \centering
        \includegraphics[width=\linewidth]{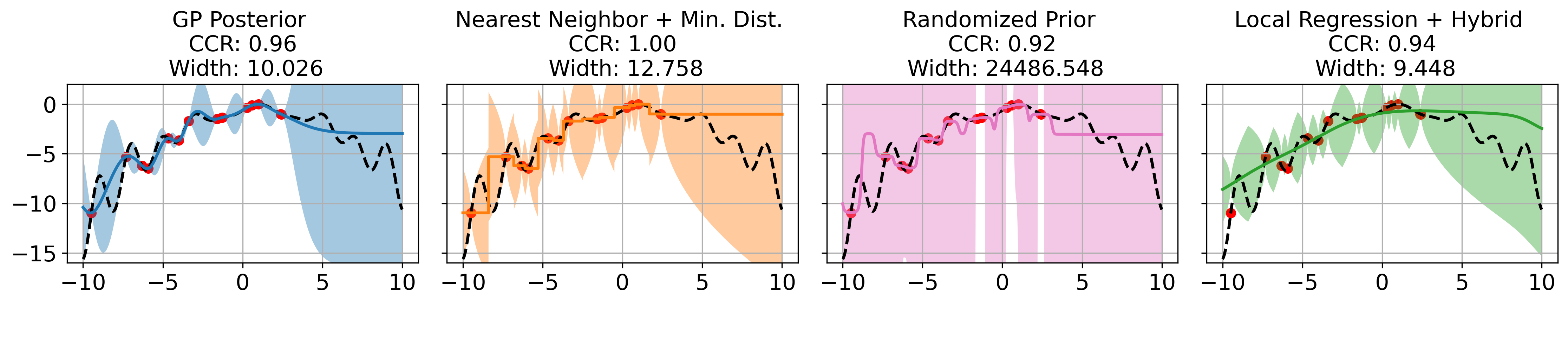}
    \end{subfigure}

    \vspace{5pt} 

    \begin{subfigure}[t]{\textwidth}
        \centering
        \includegraphics[width=\linewidth]{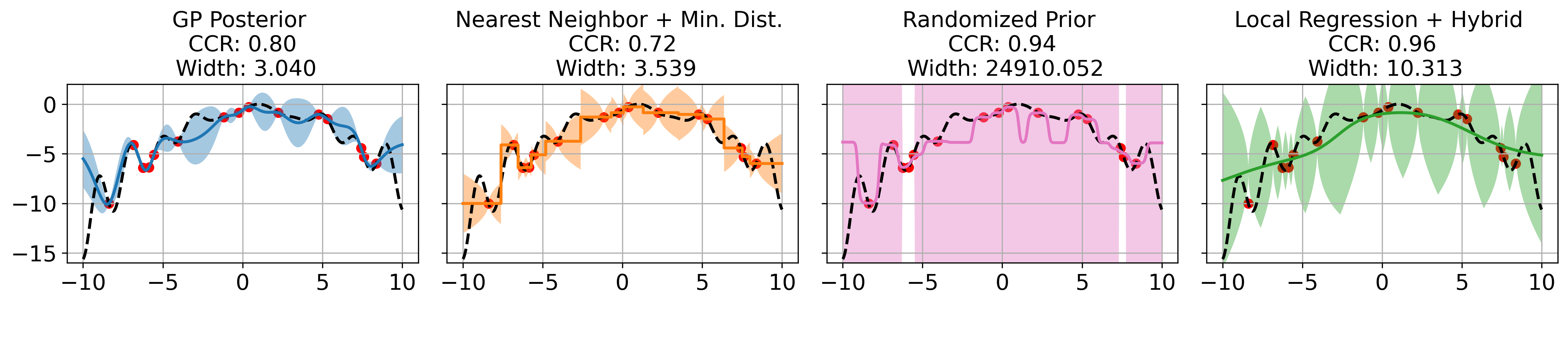}
    \end{subfigure}

    \vspace{5pt}

    \begin{subfigure}[t]{\textwidth}
        \centering
        \includegraphics[width=\linewidth]{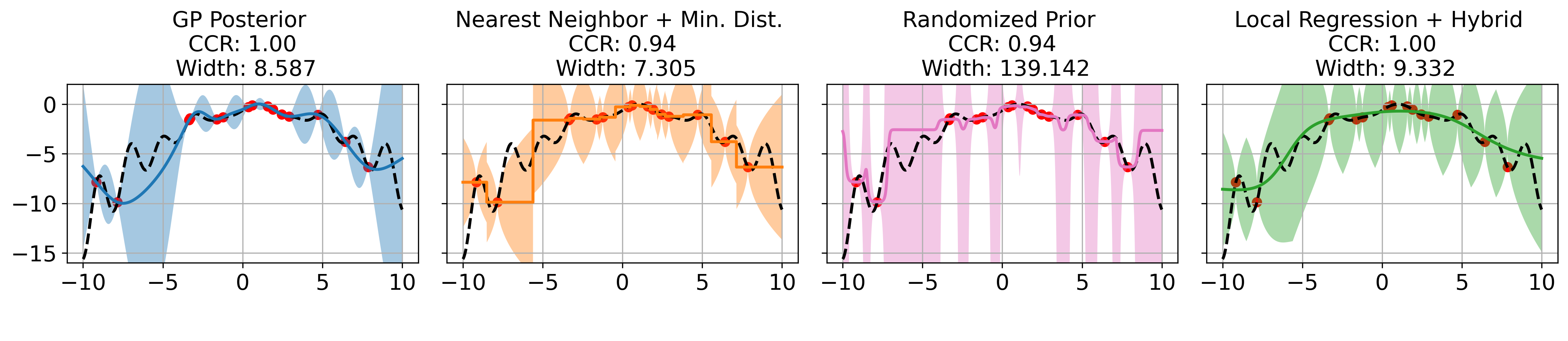}
    \end{subfigure}

    \vspace{5pt}

    \begin{subfigure}[t]{\textwidth}
        \centering
        \includegraphics[width=\linewidth]{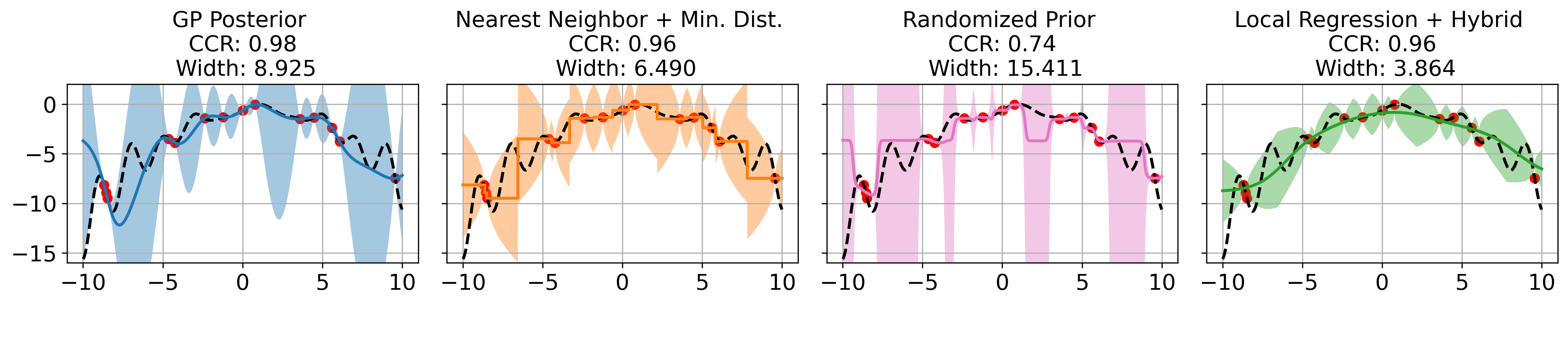}
    \end{subfigure}

    \vspace{5pt}

    \begin{subfigure}[t]{\textwidth}
        \centering
        \includegraphics[width=\linewidth]{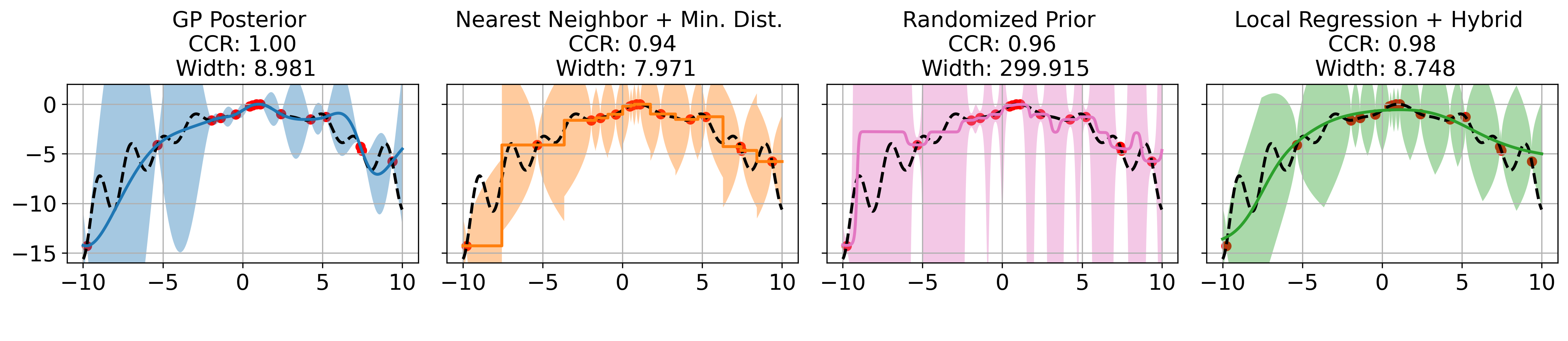}
    \end{subfigure}

    \caption{}
    \label{fig:coverage_f1}
\end{figure}

\begin{figure}[htbp!]
    \centering
    \begin{subfigure}[t]{\textwidth}
        \centering
        \includegraphics[width=\linewidth]{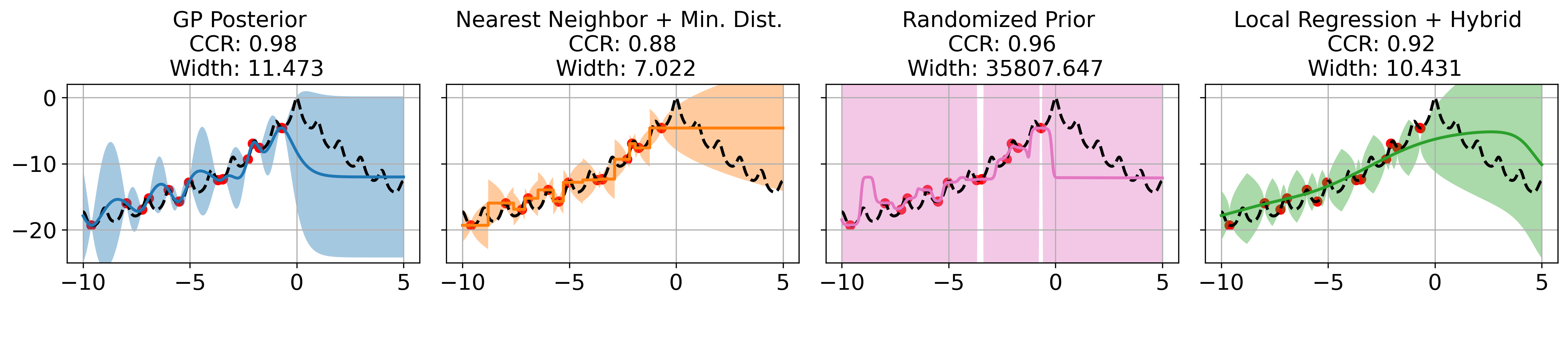}
    \end{subfigure}

    \vspace{5pt} 

    \begin{subfigure}[t]{\textwidth}
        \centering
        \includegraphics[width=\linewidth]{Figures/PseudoBO_coverage_plots/Func2_coverage_5.png}
    \end{subfigure}

    \vspace{5pt}

    \begin{subfigure}[t]{\textwidth}
        \centering
        \includegraphics[width=\linewidth]{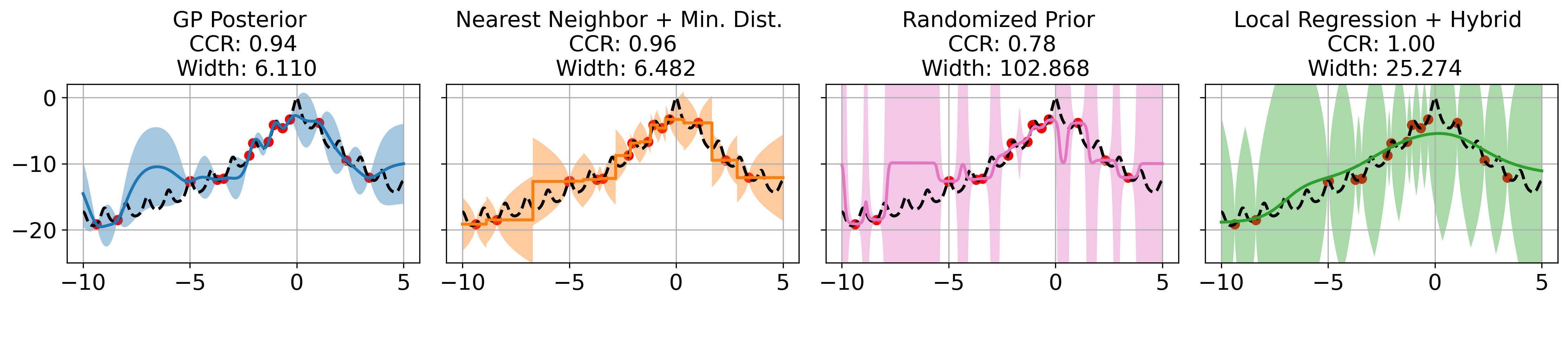}
    \end{subfigure}

    \vspace{5pt}

    \begin{subfigure}[t]{\textwidth}
        \centering
        \includegraphics[width=\linewidth]{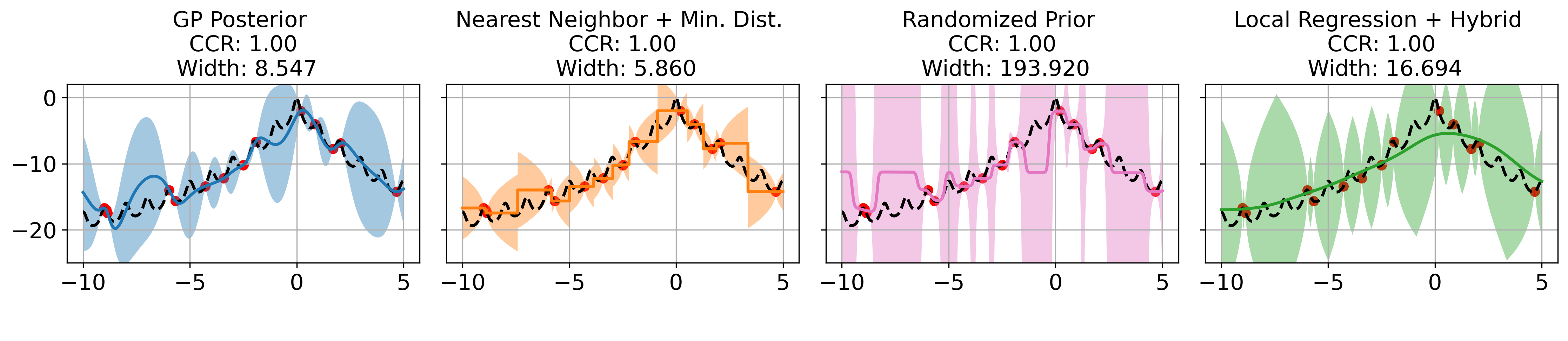}
    \end{subfigure}

    \vspace{5pt}

    \begin{subfigure}[t]{\textwidth}
        \centering
        \includegraphics[width=\linewidth]{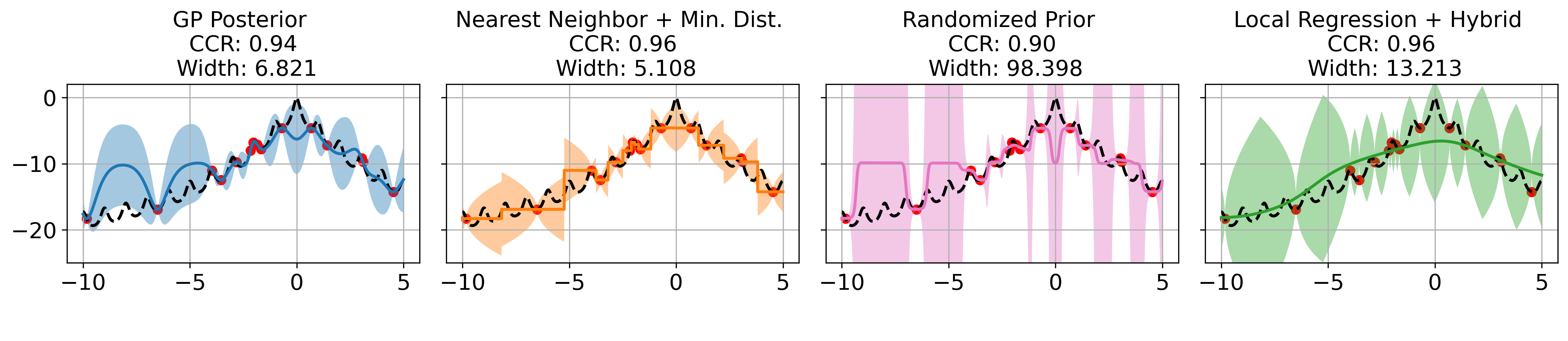}
    \end{subfigure}

    \caption{}
    \label{fig:coverage_f2}
\end{figure}

\begin{figure}[htbp!]
    \centering
    \begin{subfigure}[t]{\textwidth}
        \centering
        \includegraphics[width=\linewidth]{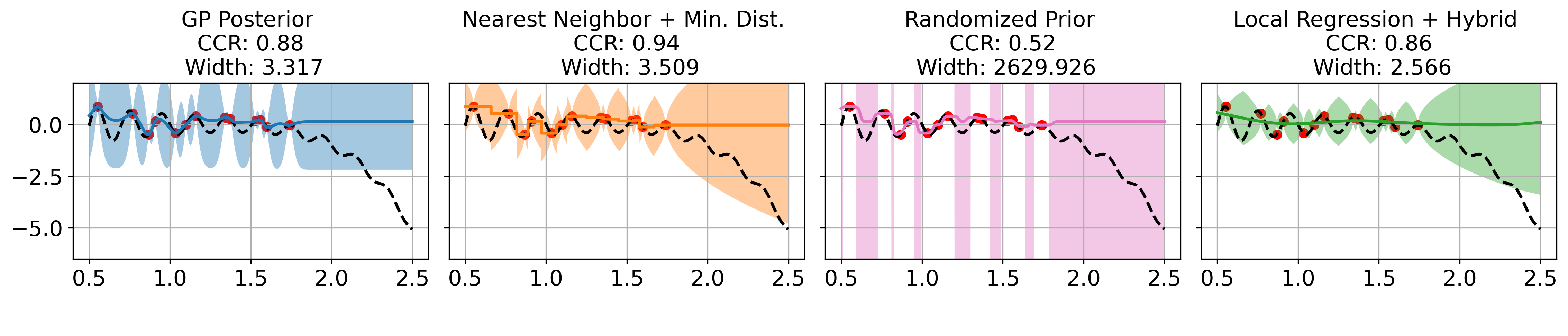}
    \end{subfigure}

    \vspace{5pt} 

    \begin{subfigure}[t]{\textwidth}
        \centering
        \includegraphics[width=\linewidth]{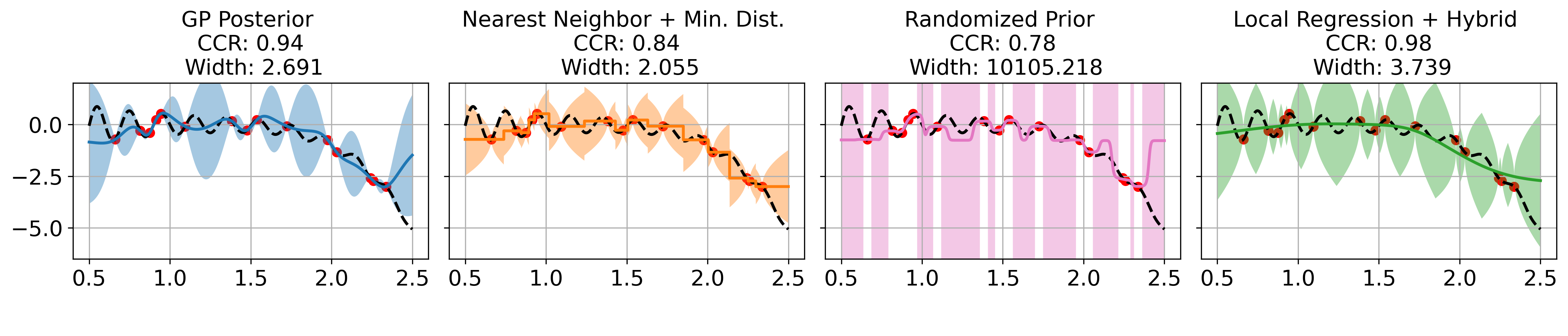}
    \end{subfigure}

    \vspace{5pt}

    \begin{subfigure}[t]{\textwidth}
        \centering
        \includegraphics[width=\linewidth]{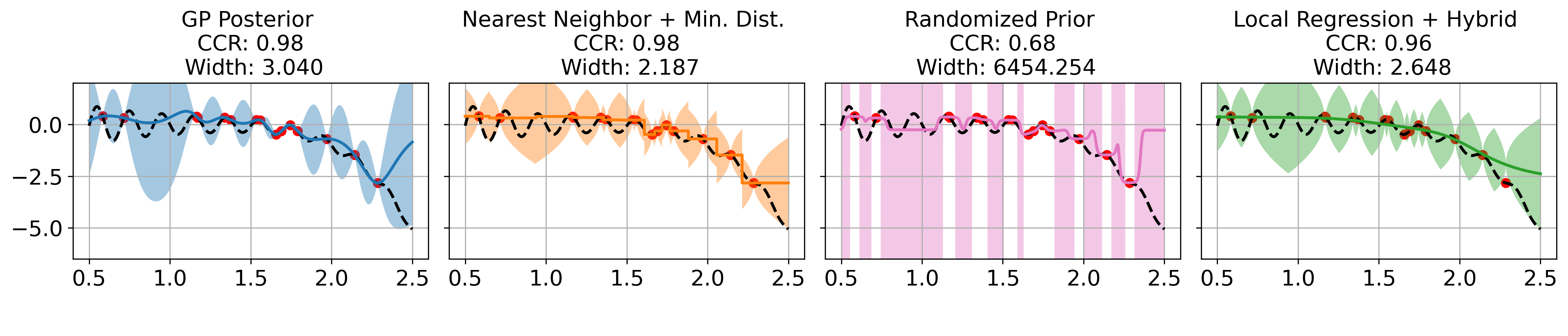}
    \end{subfigure}

    \vspace{5pt}

    \begin{subfigure}[t]{\textwidth}
        \centering
        \includegraphics[width=\linewidth]{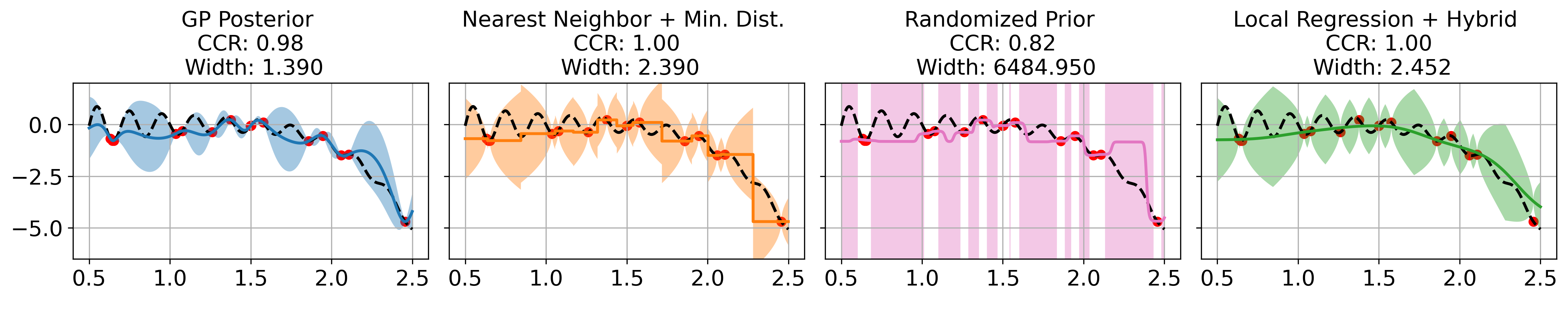}
    \end{subfigure}

    \vspace{5pt}

    \begin{subfigure}[t]{\textwidth}
        \centering
        \includegraphics[width=\linewidth]{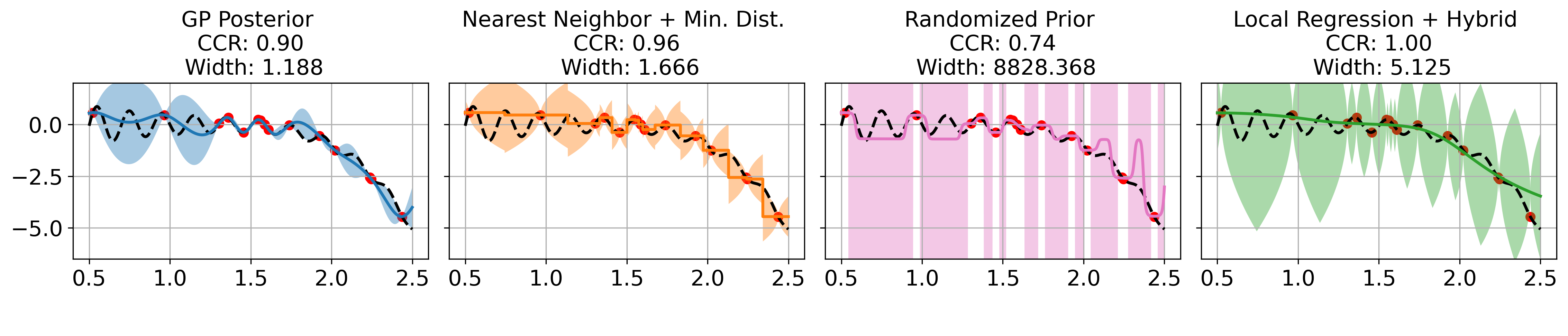}
    \end{subfigure}

    \caption{}
    \label{fig:coverage_f3}
\end{figure}

Additional sample runs on the objective $f_1$, $f_2$ and $f_3$ are shown in Figures \ref{fig:coverage_f1}, \ref{fig:coverage_f2} and \ref{fig:coverage_f3}.


\subsection{Hyperparameter Tuning}\label{subsec: tuning detail}

This task originates from HPOBench \cite{klein2019tabular}. The parameters to be tuned are shown in Table \ref{tab: HBOBench}.

\begin{table}[htbp]
  \centering
  \scalebox{1}{
  \begin{tabular}{c|c}
    \toprule
    Hyperparameter  &  Choices\\
    \midrule
    Initial LR & $\{0.0005,0.001,0.005,0.01,0.05,0.1\}$ \\
    Batch Size & $\{8,16,32,64\}$ \\
    LR Schedule & $\{$ cosine, fixed $\}$ \\
    \midrule
    Activation of Layer 1 & $\{$ Tanh, ReLU $\}$ \\
    Width of Layer 1 & $\{16,32,64,128,256,512\}$ \\
    Dropout rate of Layer 1 & $\{0.0,0.3,0.6\}$ \\
    \midrule
    Activation of Layer 2 & $\{$ Tanh, ReLU $\}$ \\
    Width of Layer 2 & $\{16,32,64,128,256,512\}$ \\
    Dropout rate of Layer 2 & $\{0.0,0.3,0.6\}$ \\
    \bottomrule
  \end{tabular}}
  \caption{\small{Tunable hyperparameters and search space in the HBOBench task.}}\label{tab: HBOBench}
\end{table}

In this task, we follow the procedure in \cite{tiao2021bore} to densify the search space by one-hot encoding for categorical hyperparameters (including LR Schedule, activation function of layers $1$ and $2$), with each variable with domain [0,1] representing each category. For each of the other hyperparameters, we use one variable with domain [0,1] with space equalized separated for all candidates of that hyperparameter. Therefore, the search domain is $[0,1]^{12}$. 

\subsection{Robot Arm Pushing}
This 14D task is designed for optimizing the controllers of the robot's two arms to push two objects to their target places $o_1$ and $o_2$, starting from positions $s_1$ and $s_2$. Let us denote the ending positions of the two objects by $e_1$ and $e_2$, resulted from a specific control. The final reward is defined as
\begin{equation*}
    R(x) = \norm{s_1 - o_1} + \norm{s_2 - o_2} - (\norm{s_1 - e_1} + \norm{s_1 - e_2}).
\end{equation*}
There are $14$ parameters to control the location and rotation of the robot hands, pushing speed, moving direction and pushing time, presented in Table \ref{tab: RL14D}.

\begin{table}[htbp]
  \centering
  \scalebox{1}{
  \begin{tabular}{c|c}
    \toprule
    Hyperparameter of Arm 1/2 &  Ranges \\
    \midrule
    Position $x$ & $[-5, 5]$ \\
    Position $y$ & $[-5, 5]$ \\
    Angle & $[0, 2\pi]$ \\
    Torque & $[-5, 5]$ \\
    Velocity $v_x$ & $[-10, 10]$ \\
    Velocity $v_y$ & $[-10, 10]$ \\
    Push duration & $[2, 30]$\\
    \bottomrule
  \end{tabular}}
  \caption{\small{Tunable hyperparameters and search space in the robot arm pushing task.}}\label{tab: RL14D}
\end{table}


\subsection{Rover Trajectory Planning}
This task is a $60$D problem, where our target is to optimize the trajectory of the rover, determined by our choices of $30$ points, in a 2D plane. The reward is estimated in the following way:
\begin{equation*}
    f(x) = c(x) + \lambda(\norm{x_{0,1} - s}_{1} + \norm{x_{59,60} - o}_{1}) + b,
\end{equation*}
where $s$ and $o$ are the starting position and the target position, $x \in [0, 1]^{60}$ containing the points picked, and $c(x)$ is a function to measure the cost of the trajectory determined by $x$.

Instructions for running these two tasks can be found in \href{https://github.com/zi-w/Ensemble-Bayesian-Optimization}{https://github.com/zi-w/Ensemble-Bayesian-Optimization}.

\end{document}